\PassOptionsToPackage{no-math}{fontspec}
\documentclass[9pt,a4paper,twocolumn,twoside]{rho-class/rho}
\usepackage[english]{babel}

\usepackage{xcolor}
\newcommand{\lj}{LLM Jury~}
\newcommand{\LJ}{LLM Jury~}

\usepackage{makecell}
\usepackage{placeins}

\usepackage{tabularray}
\usepackage{booktabs}
\usepackage{siunitx}
\makeatletter
\DeclareSizeFunction{sub}{\sub@sfcnt\@font@info}
\makeatother

\usepackage{mathtools}
\newcommand{\stackci}[3]{%
  \ensuremath{%
    #1^{\mathmakebox[2.3em][r]{#3}}_{\mathmakebox[2.3em][r]{#2}}%
  }%
}
\usepackage{anyfontsize}

\usepackage{bm}

\title{Can LLMs Accurately Score Medical Diagnoses and Clinical \\Reasoning?
}

\author[{\textasteriskcentered}, a]{Amy Rouillard}
\author[b,c]{Sitwala Mundia}
\author[b]{Linda Camara}
\author[d]{Ziyaad Dangor}
\author[d]{Michael Cameron Gramanie}
\author[a,e,f]{Ismail Kalla}
\author[d]{Shabir A. Madhi}
\author[d]{Kajal Morar}
\author[d]{Marlvin T. Ncube}
\author[g]{Haroon Saloojee}
\author[{\textasteriskcentered}, a,b,h]{Bruce A. Bassett}

\affil[a]{Wits MIND Institute, University of the Witwatersrand, Johannesburg, South Africa}
\affil[b]{Grai Labs, Cape Town, South Africa}
\affil[c]{Faculty of Health Sciences, University of the Witwatersrand, Johannesburg, South Africa}
\affil[d]{
South African Medical Research Council Vaccines and Infectious Diseases Analytics Research Unit, Faculty of Health Sciences, University of the Witwatersrand, Johannesburg, South Africa}
\affil[e]{Department of Internal Medicine, Charlotte Maxeke Johannesburg Academic Hospital, Johannesburg, South Africa}
\affil[f]{School of Clinical Medicine, University of the Witwatersrand, Johannesburg, South Africa}
\affil[g]{Department of Paediatrics and Child Health, Faculty of Health Sciences, University of the Witwatersrand, Johannesburg, South Africa}
\affil[h]{School of Computer Science and Applied Mathematics, University of the Witwatersrand, Johannesburg, South Africa}

\corres{\textsuperscript{\textasteriskcentered} Corresponding authors: \{amy.rouillard,bruce.bassett1\}@wits.ac.za }

\begin{abstract}
    Evaluating medical AI systems using expert clinician panels is costly and slow, motivating the use of large language models (LLMs) as alternative adjudicators. Here, we evaluate an LLM Jury, composed of three frontier AI models, for scoring 3334 diagnoses on 300 real-world low- and middle-income country (LMIC) hospital cases. Both LLM- and clinician-generated diagnoses are scored against expert panel diagnoses across four dimensions: diagnosis, differential diagnosis, clinical reasoning, and negative treatment risk. The LLM Jury scores are compared with expert and independent re-scoring panel scores to assess error metrics, inter-rater agreement, severe-risk errors, and the effect of post hoc calibration using isotonic regression. In our data, we find that: (i) the uncalibrated LLM Jury scores preserve ordinal agreement with the expert clinician panel scores, but are systematically lower; (ii) the probability of severe-risk errors is lower for the LLM Jury than the human expert re-score panels; (iii) the LLM Jury combined with LLM diagnoses can be used to identify diagnoses at high risk of error, enabling targeted expert review and improved panel efficiency; (iv) the calibrated LLM Jury scores and rankings of diagnosing agents show excellent agreement with those of the primary expert panels; (v) LLM Jury models show no self-preference bias, they did not score diagnoses generated by their own underlying model or models from the same vendor more (or less) favourably than those generated by other models. Together, these results provide evidence that a calibrated LLM Jury is a trustworthy and reliable proxy for expert clinician evaluation in medical AI benchmarking. Confirming these findings in other clinical settings is an important direction for future work.
\end{abstract}

\begin{document}

    \maketitle
    \thispagestyle{firststyle}

\section{Introduction}

\rhostart{H}ealthcare applications of generative Artificial Intelligence (AI) demand rigorous model evaluation to ensure safety and accuracy. Traditionally, expert clinicians review AI-generated answers, but this approach is labour-intensive, inconsistent, and not easily scalable~\cite{williams_human_2025,arora_healthbench_2025,bedi_holistic_2026}. Evaluating a number of AI models rigorously across a large number of cases with human panels is simply not feasible. Researchers have instead explored using large language models (LLMs) as evaluators, an approach often termed “LLM-as-a-Judge”, to provide efficient and reproducible assessments of medical AI outputs~\cite{genovese_artificial_2026,williams_human_2025,croxford_automating_2025,li_generation_2025}. Early results suggest that strong LLM-based judges can approximate human judgment closely~\cite{chiang_can_2023,zheng_judging_2023}. Such findings indicate that LLM evaluators could serve as a scalable proxy for human review in medicine, where obtaining consistent expert feedback is often costly and slow.

While LLM-as-judge frameworks offer scalability, relying on a single model as the sole evaluator presents notable limitations. Individual LLM judges may exhibit systematic provider-specific biases, favour lengthier responses or struggle with nuanced clinical reasoning~\cite{bedi_holistic_2026}. Although the reliability of automated grading can be enhanced through highly granular, case-specific rubrics~\cite{arora_healthbench_2025}. This approach is constrained by the high manual effort required to develop such rubrics for diverse clinical scenarios. Furthermore, in a recent assessment of clinical support in Rwanda, even best-performing LLM judges achieved expert-level parity on only $4$ of $11$ evaluation criteria, with some models proving consistently overcritical while others were overly lenient~\cite{williams_human_2025}. Crucially, these automated judges showed significantly degraded performance when evaluating non-English inputs such as Kinyarwanda and failed to identify demographic biases spotted by local doctors.

In light of the above trends, our paper assesses a frontier-level \LJ for evaluating medical AI outputs with expert-level rigour in a low- and middle-income country (LMIC) context. While prior studies have tested multi-LLM evaluators in general domains and even shown modest gains in specialised medical settings~\cite{williams_human_2025,croxford_evaluating_2025,reese_using_2026}, the approach has not yet been fully validated for clinical AI applications. We fill this gap by systematically assessing how a panel of LLM judges can reproduce expert panel scoring on medical diagnosis assessment. In our data, we found that: 
\begin{itemize}
    \item The uncalibrated LLM Jury scores are systematically lower than the primary expert panels.
    \item The LLM Jury preserves ordinal agreement with the primary expert panels.
    \item The probability of severe-risk errors, instances where the primary human expert panel identified a diagnosis as posing significant risk to the patient, but another evaluator did not, was significantly lower in \lj models compared to independent human expert re-score panels.
    \item The LLM Jury combined with LLM diagnoses can be used to identify diagnoses at high risk of error, enabling targeted expert review and improved panel efficiency.
    \item The LLM Jury rankings of diagnosing agents show excellent agreement with those of the primary human expert panels.
    \item The calibrated LLM Jury matches or outperforms the human expert re-score panels across all metrics.
    \item The models making up the LLM Jury show no self-preference bias. They did not score diagnoses generated by their own underlying model or models from the same vendor more (or less) favourably than those generated by other models.
\end{itemize}
These results indicate that a suitable \lj can indeed serve as a useful and trustworthy surrogate for human evaluation in our setting. An important direction for future work is to validate these findings in other clinical contexts.

A shorter version of this work was presented at Artificial Intelligence in Medicine~\cite{Rouillard2026LLMJury}. Compared to the conference version, this paper provides an expanded analysis of the LLM Jury's reliability and stability. Specifically, we include a detailed examination of severe-risk errors (Section~\ref{sec:sre}), an assessment of scoring stability across multiple model iterations (Section~\ref{sec:score-variablity}), and an evaluation of the LLM Jury's ability to flag incorrect ward diagnoses (Section 7). This version also includes further details on post hoc calibration of LLM scores using isotonic regression (Section~\ref{sec:cal}) and provides an extensive appendix with full confusion matrices, cross-validation curves, and expanded inter-rater reliability metrics.

\section{Data, Panels and \lj}\label{sec:data-and-panels}

{\bf Datasets} used in this work were collected as part of the VALID study~\cite{bassett_multimodal_2026}. The VALID study evaluated the multimodal diagnostic performance of LLMs on a South African public health hospital cohort. During the study, a team of medical officers captured $539$ complete medical cases together with primary, secondary, and differential diagnoses made by the treating physicians, which we refer to collectively as the {\em ward diagnoses}. This multimodal dataset comprises admission records, structured clinical reports, laboratory test results, and medical imaging data, including computerised tomography (CT), magnetic resonance imaging (MRI) and X-rays. The dataset is not publicly available, making it suitable for independent evaluation of LLMs.

{\bf Primary panel diagnoses} are diagnoses made by expert panels of two internal medicine physicians or two paediatricians. These panels reviewed 300 speciality-relevant cases: members examined cases independently in advance, then met to provide consensus primary, secondary, and differential diagnoses, document clinical reasoning, and record binary agreement with ward diagnoses (examples in \cite{bassett_multimodal_2026}). {\bf Primary panel scores} are evaluations provided by the primary panels. In total, they scored 30 ward diagnoses and 300 LLM diagnoses, one per case, against their own expert diagnoses. This yielded four Likert-scale scores (1–5): diagnosis (Dx), differential diagnosis (DDx), clinical reasoning (Reasoning), and negative treatment risk. Higher scores are better except for negative treatment risk, which we replace with a patient safety (Safety) score, defined as 6 minus the negative treatment risk. The clinical reasoning of the treating doctors was not available; therefore, the clinical reasoning score is omitted for ward diagnoses. 

{\bf The LLM Jury} was formed from three leading LLM models: Anthropic's Claude Opus 4.1 (2025-08-05), Google's Gemini 2.5 Pro (2025-06-17) and OpenAI's o3 (2025-04-16). These reasoning AI models were chosen from different labs to enhance diversity. The \lj models were identically prompted to provide each of the four scores (Dx, DDx, Reasoning, negative treatment risk) on the same 1-5 scale as the human panels. In total, each \lj model evaluated $3334$ diagnoses: $300$ ward diagnoses associated with the cases reviewed by the primary panels, and $3034$ LLM diagnoses made by a variety of models. In the \lj prompt, the primary panel's diagnoses and clinical reasoning were supplied as ground truth. The final \LJ scores are computed as the mean of the three \lj model scores. We do not round the \LJ scores to integer values, preserving the granularity of the aggregated evaluations and reflecting uncertainty arising from disagreement among the jury models.

The \textbf{re-score panels}, a second set of expert panels, re-scored a random duplicate sample of $30$ LLM diagnoses to assess inter-panel agreement. Re-score panels were not provided with the full case details, only the AI model predictions and the primary panel's diagnoses and clinical reasoning as ground truth, exactly mirroring the data provided to the LLM Jury. For efficiency, the primary panels scored diagnoses immediately after recording their diagnoses and clinical reasoning, and therefore were aware of the full case details. However, because scoring was anchored to concordance with the panel's own diagnoses and reasoning using predefined criteria, this context was unlikely to systematically bias the assigned scores, although this should be explored in future work.

Table~\ref{tab:num-dx-eval} provides a detailed summary of the number of diagnoses evaluated, categorised by the diagnosing entity. Table~\ref{tab:num-dx-eval} also splits the case diagnoses into those used for calibration, discussed here, and those used for diagnostic evaluation, presented in \cite{bassett_multimodal_2026}. Human expert evaluation took place throughout the course of the VALID study, and as a result, the 23 LLMs evaluated in the calibration sample encompass a broad range of models. This diversity ensures a wide range of diagnostic quality, allowing for robust calibration of the LLM Jury. The final evaluation utilised 10 LLMs from 4 vendors, representing the most current model versions available at the time. This work utilises the calibration sample to assess the quality of the LLM Jury. The results obtained from the evaluation sample ($n=3034$), which is an order of magnitude larger than the calibration sample ($n=330$), are reported in \cite{bassett_multimodal_2026}.

\section{Summary Statistics: Human Panels vs LLM Jury Scores}\label{sec:summary-stats}

In this section, we present summary statistics comparing the scores assigned by the primary panels with those assigned by the LLM Jury and the re-score panels. We assess the performance of each \lj model individually as well as the performance of the LLM Jury, which refers to the mean score over the three \lj models. Across all scores, the primary panels and re-score panels assigned, on average, higher scores (mean $3.0$-$3.7$; median $3$-$4$; Figure~\ref{fig:score_distributions}) than the \lj models (mean $2.5$-$3.3$; median $2$-$4$; Figure~\ref{fig:score_distributions}).

Analysis of the score differences between the primary panels and the other evaluators, Table~\ref{tab:offset_rmse}, indicates that all \lj models systematically penalised diagnostic quality relative to the panels' scores. Compared to the other \lj models, o3 has the lowest offset and root mean squared error (RMSE). The re-score panels consistently have the smallest offsets compared to all other evaluators, while their RMSEs were comparable to the \lj models. Table~\ref{tab:offset_rmse} also provides the 95\% confidence intervals (CI) for the offset and RMSE, respectively, calculated for each evaluator with respect to the primary panel. As a result of the difference in sample size (30 vs 330 cases), the 95\% confidence intervals are wider for the re-score panels than for the \lj models. All \lj models have a similar offset and RMSE for the differential diagnosis score. For other scores, there are larger differences in evaluator performance and, among the \lj models, the lowest offset and RMSE is achieved by o3.

\begin{figure*}[htb]
    \centering
    \begin{subfigure}[t]{0.45\linewidth}
        \centering
        \includegraphics[width=\linewidth]{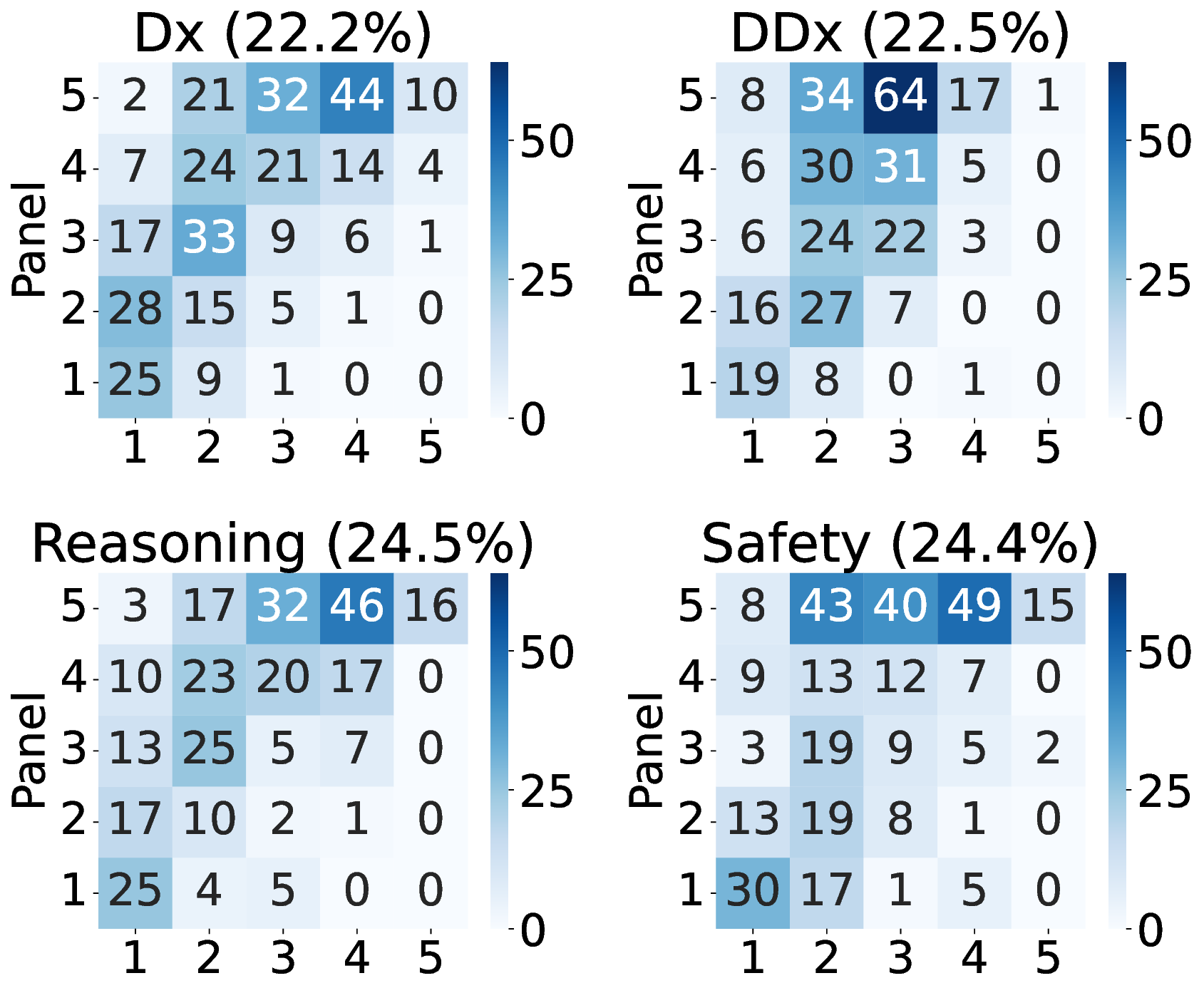}
        \caption{Opus 4.1 
        }
        \label{fig:CM_anthropic_claude-opus-4-1}
    \end{subfigure}
    \hspace{0.5em}
    \begin{subfigure}[t]{0.45\linewidth}
        \centering
        \includegraphics[width=\linewidth]{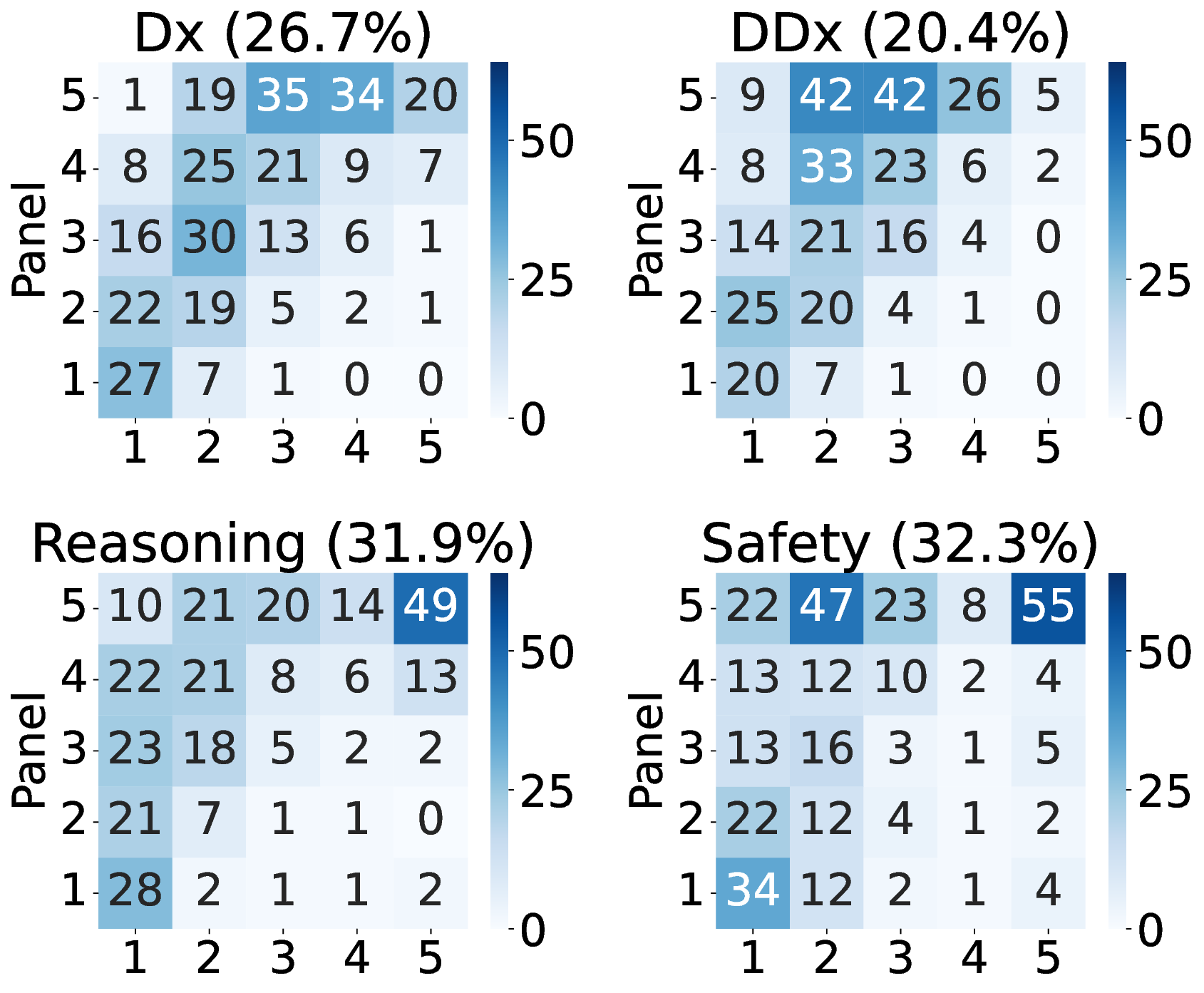}
        \caption{Gemini 2.5 pro 
        }
        \label{fig:CM_gemini-2.5-pro}
    \end{subfigure}
    \vspace{0.5em}

    \begin{subfigure}[t]{0.45\linewidth}
        \centering
        \includegraphics[width=\linewidth]{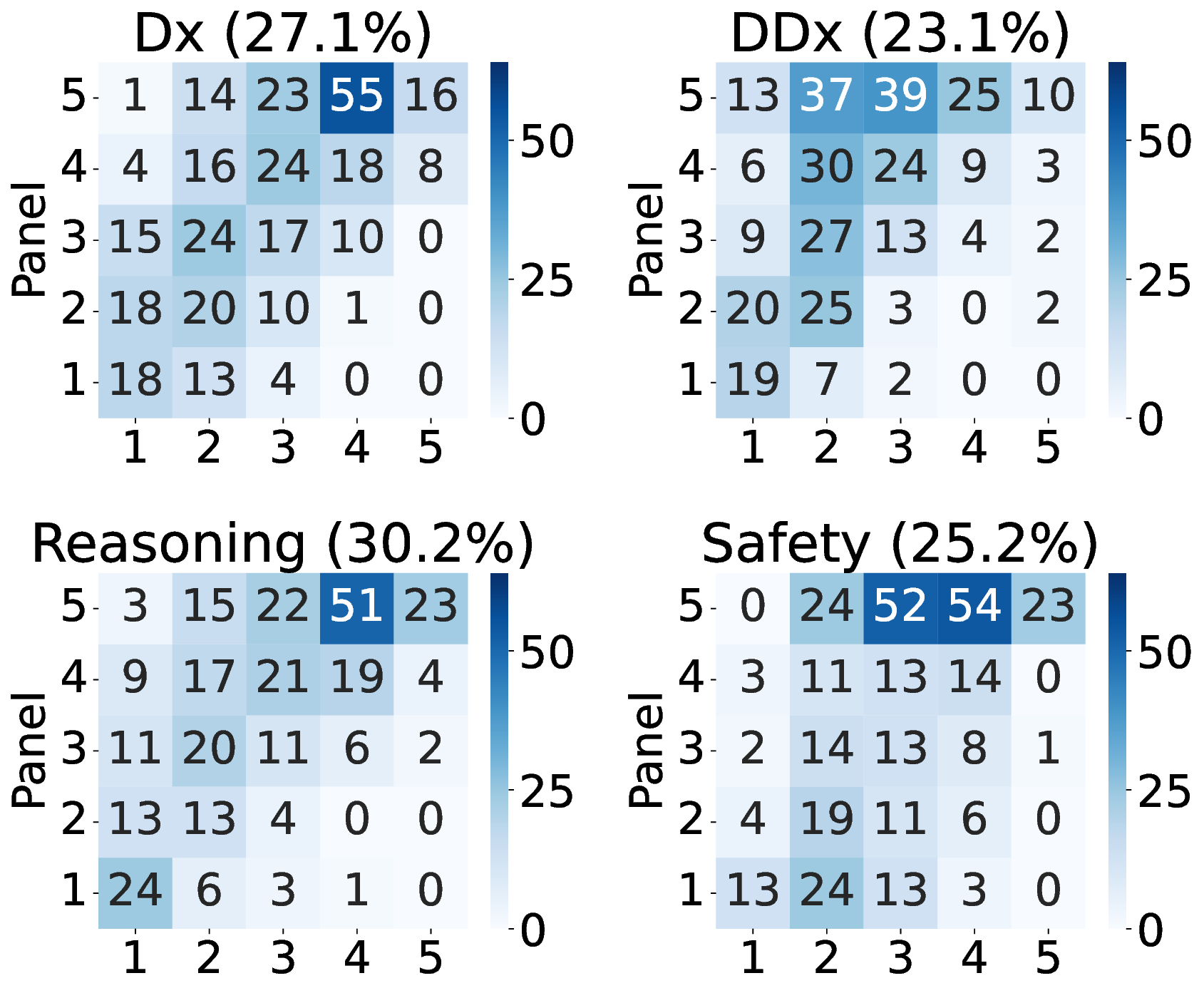}
        \caption{o3 
        }
        \label{fig:CM_o3-2025-04-16}
    \end{subfigure}
    \hspace{0.5em}
    \begin{subfigure}[t]{0.45\linewidth}
        \centering
        \includegraphics[width=\linewidth]{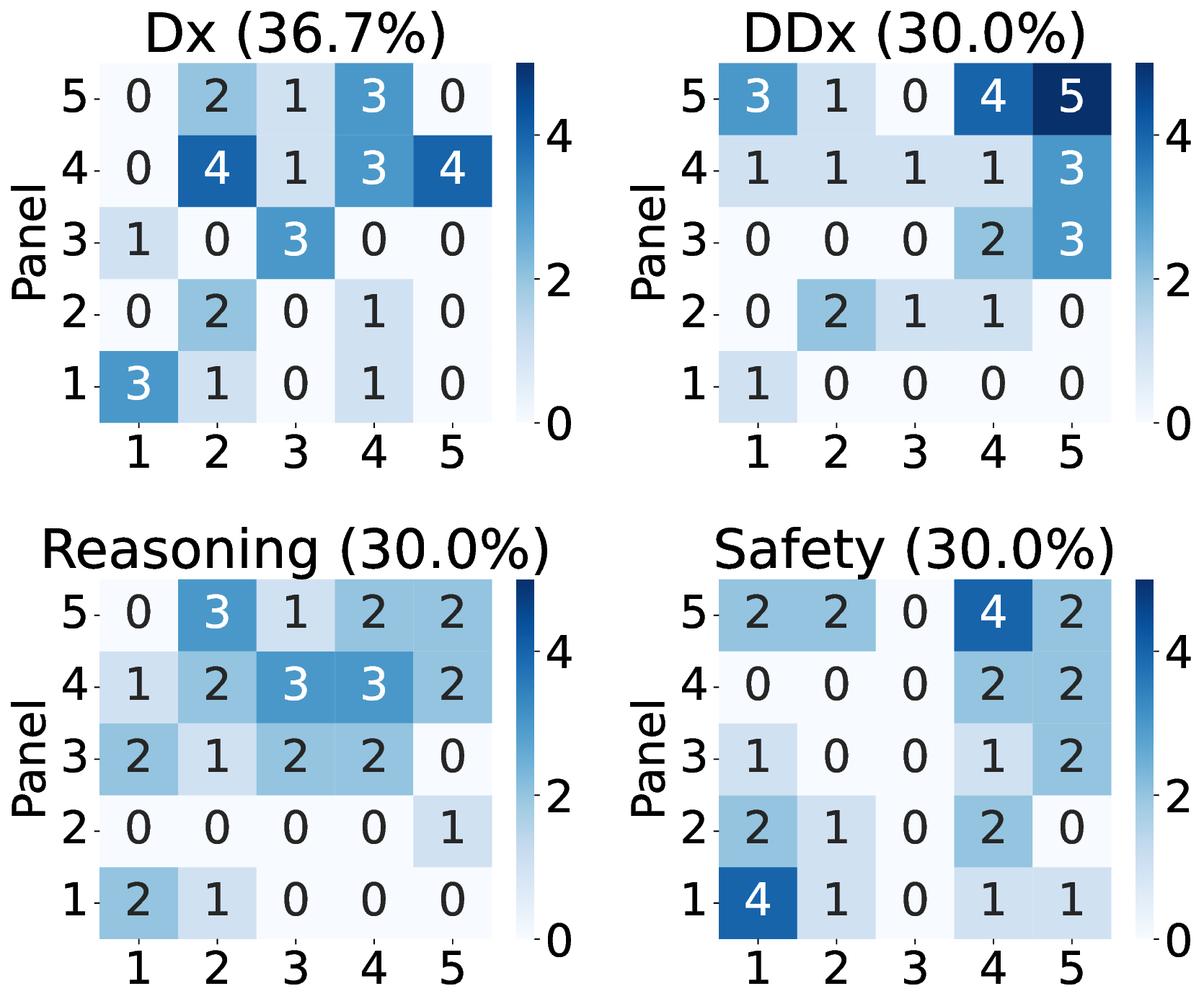}
        \caption{Re-score Panels 
        }
        \label{fig:CM_panel-rescore}
    \end{subfigure}
    \caption{\textbf{Scoring comparison.} Confusion matrices showing the concordance between the primary panels and (a)-(c) the \lj models and (d) the re-score panels for each of the four scores. The percentage of exact matches is shown in brackets. The \lj models (a)-(c) systematically assign lower scores compared to the primary panels, while the re-score panels (d) differ symmetrically from the primary panels.   
    } 
    \label{fig:CMs}
\end{figure*}

Figure~\ref{fig:CMs} shows confusion matrices allowing comparison of the scores given by the expert panels to the scores given by the \lj models and the re-score panels. These confusion matrices demonstrate the consistent strictness of the \lj models (most data lie above the diagonals). In particular, the \lj rarely gives the highest score, 5. The exact match percentage across all scores for the \lj is low (20\%-32\%), indicating that they often disagreed with the primary panels on the precise rating. However, the ordinal consistency is generally preserved. In contrast, the human re-score panels (bottom right) had more random, symmetric disagreement with the primary panels, with lower correlations and higher exact match percentages (30\%-37\%).

\section{Panel-LLM Jury Inter-rater Agreement }\label{sec:inter-rater-agreement}

To evaluate the consistency of the judgments provided by the LLM Jury and re-score panels relative to the primary panels, we calculate two complementary inter-rater agreement metrics. In Table~\ref{tab:rho_kappa}, we report the agreement between each evaluator and the primary panel using: Spearman's $\rho$, defined as the Pearson correlation coefficient between rank variables, which is a non-parametric measure of agreement of orderings and is insensitive to calibration errors~\cite{spearman_proof_1987}; And, quadratic weighted Cohen’s $\kappa$, which quantifies absolute agreement and allows for weighting of ordinal differences~\cite{cohen_coefficient_1960,fleiss_equivalence_1973}. There is fair to moderate agreement between the \lj and the primary panels, with consistently better agreement than between the re-score and the primary panels. We note that the diagnoses evaluated by the re-score panels represent a small subset of those evaluated by the LLM Jury, and the observed differences are not all statistically significant. Across all \lj models and all metrics, inter-rater agreement is highest for the diagnosis score and lowest for the differential diagnosis score. Notably, the \LJ has the highest Spearman’s $\rho$ and o3 the highest Cohen's $\kappa$ compared to all other evaluators, including the re-score panels.

\begin{figure*}[htbp]
  \centering
  \includegraphics[width=0.9\linewidth]{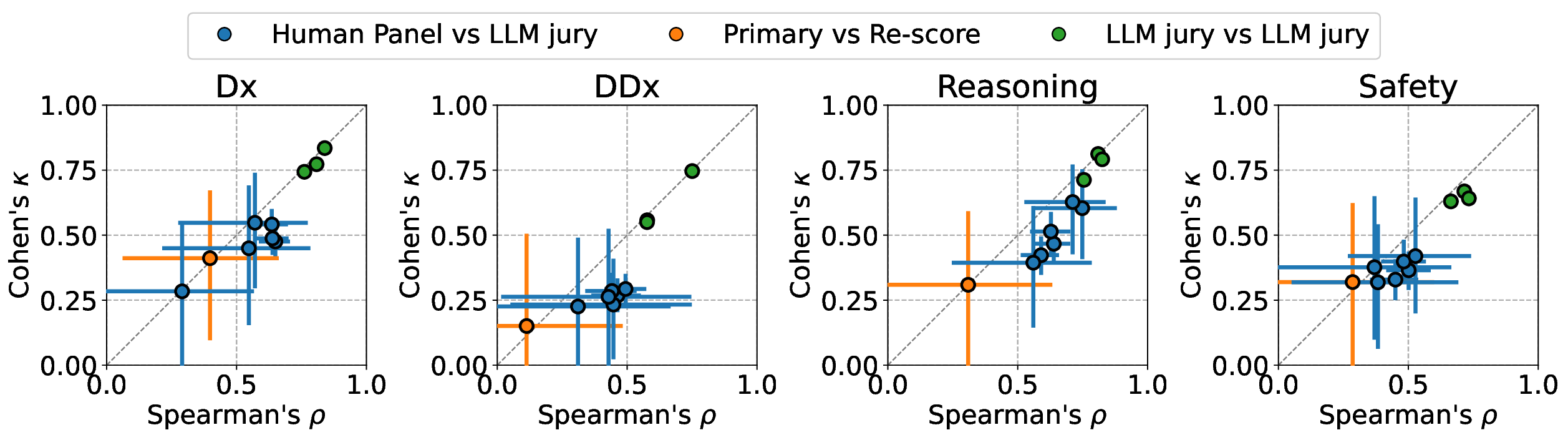}
  \caption{\textbf{Inter-rater agreement.} Spearman's $\rho$ versus quadratic weighted Cohen's $\kappa$ where face colour indicates which evaluators are being compared. Error bars reflect 68\% confidence intervals (CI) computed using bootstrap sampling ($n =1000$). The LLM Jury-human panel agreement (blue) is better than the human-human panel (primary or re-score) inter-rater agreement (orange). However, the blue points lie slightly below the diagonal, indicating weaker exact agreement than rank-order consistency between the \lj and human panels. The inter-rater agreement between \lj models (green) is greater than both human-human agreement and human-\lj agreement, showing the high degree of coherence between the \lj models.
  }\label{fig:spearman-vs-cohen-2-by-2}
\end{figure*}

Figure~\ref{fig:spearman-vs-cohen-2-by-2} shows Spearman's $\rho$ versus Cohen's $\kappa$ between the two evaluator groups, human and LLM. Across all scores, there is a strong positive association between metrics. However, Spearman’s $\rho$ is marginally higher than weighted Cohen’s $\kappa$ for the human panels vs \lj models (blue), indicating greater ordinal alignment than absolute agreement. This is consistent with the observation that the \lj models give, on average, lower scores compared to the primary panels, Section~\ref{sec:summary-stats}. Across all scores except for the diagnosis score, the primary vs the re-score panels (orange) exhibit the poorest agreement compared to any other evaluator pairs, consistent with Figure~\ref{fig:CMs}. For the diagnosis score, the worst inter-rater agreement was between the re-score panels and Gemini 2.5 Pro. The agreement between the \lj models (green) is high across all scores. Gemini 2.5 Pro and Opus 4.1 had the lowest overall inter-rater agreement, while the highest overall inter-rater agreement was between o3 and either Gemini 2.5 Pro or Opus 4.1.

\subsection{Do the LLM Jury Model Reason Consistently?}

By examining the correlations among scores assigned by each evaluator, we can assess the internal consistency of the \lj models. 
The primary panels demonstrate strong pairwise correlations among the diagnosis, clinical reasoning, and patient safety scores ($0.68$-$0.77$; Figure~\ref{fig:inter_score_correlation}), whereas the differential diagnosis score has weak correlations with the other scores ($0.09$-$0.52$; Figure~\ref{fig:inter_score_correlation}). This reflects the fact that differential diagnoses are formulated solely based on the patient’s initial presentation, before the availability of diagnostic test results or treatment response. Consequently, the differential diagnosis score should be assessed independently. Importantly, this pattern is also observed in the \lj models, where the pairwise correlations between the diagnosis, clinical reasoning, and patient safety scores were high ($0.73$-$0.88$; Figure~\ref{fig:inter_score_correlation}), and the differential diagnosis score is weakly correlated with all other scores ($0.09$-$0.34$; Figure~\ref{fig:inter_score_correlation}).

\section{Severe-risk Errors: \lj vs Human Evaluators}\label{sec:sre}

While the \lj scores are systematically lower than the primary panels', this does not exclude the possibility of the \lj misclassifying a subset of dangerous misdiagnoses as harmless. To assess this risk, we define the \textit{severe error rate} as the proportion of potentially harmful diagnoses (primary panel's patient safety score $\leq 2$) that also received a patient safety score from an \lj model or re-score panel of at least $3$ points higher (i.e. diagnoses with strong disagreement with the primary panel on potential harm). This corresponds to the error rate for the binary classification of `patient risk', which can be computed from Figure~\ref{fig:CMs}.  Table~\ref{tab:severe_rates} shows the severe error rates for the \lj models (approx $5\%$), which are significantly lower than the severe error rate estimated for the re-score panels ($16.7 \%$). In a beta–binomial model with a flat prior, the posterior probability that the severe error rate in the \lj is lower than that of the re-score panel is $\simeq 96.3\%$. These estimates provide further confidence in the quality of the LLM Jury.  

\begin{figure}[hbtp]
  \centering
  \includegraphics[width=0.75\linewidth]{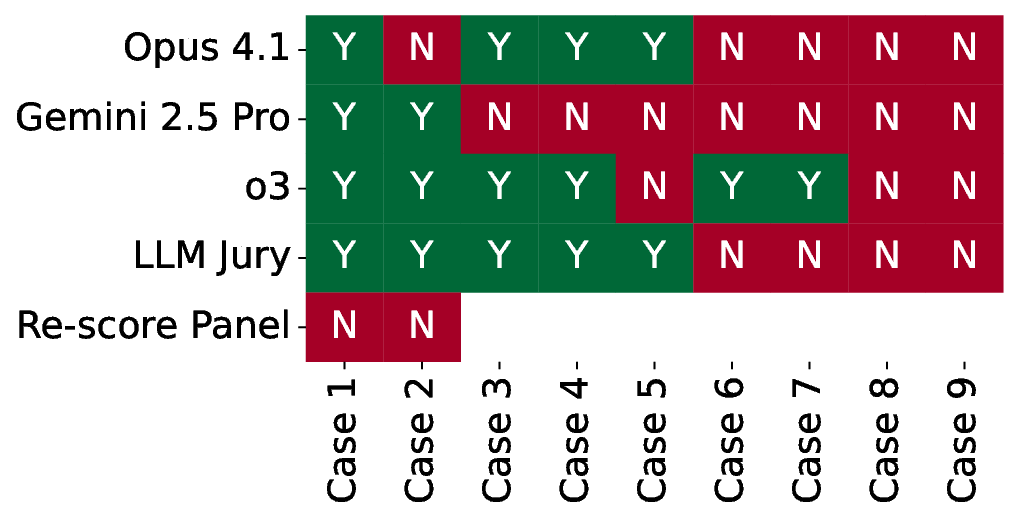}
  \caption{\textbf{Severe-risk errors.} Cases where diagnoses were evaluated by the primary panel as high risk to the patient (patient safety score of 1 or 2) while at least one other evaluator disagreed (patient safety score higher by 3 or more). Red N (green Y) indicates that the evaluator disagreed (agreed) with the primary panel. Of cases 1-9, only cases 1 and 2 were evaluated by the re-score panel. }\label{fig:sever_case_overlap}
\end{figure}

The \lj model scores are highly correlated ($0.57$-$0.76$; Figure~\ref{fig:spearman-vs-cohen-2-by-2}). As a result, the aggregated \LJ does not achieve a lower severe error rate than the individual \lj models. As shown in Figure~\ref{fig:sever_case_overlap}, $5$ out of $8$ severe risk assessment errors made by at least one \lj model were made by a majority of \lj models, resulting in \LJ having a severe risk assessment error in $4$ out of $8$ cases. Given the limited number of cases, we cannot draw strong conclusions about the underlying failure modes. Despite this limitation, a qualitative examination of individual cases can provide useful insights. Here, we highlight a few specific examples.

\begin{table}[hbtp]
\centering
\renewcommand{\arraystretch}{1.2}
\begin{tabular}{
p{0.07\linewidth}
  >{\centering\arraybackslash}p{0.09\linewidth}
  >{\centering\arraybackslash}p{0.1\linewidth}
  >{\centering\arraybackslash}p{0.09\linewidth}
  >{\centering\arraybackslash}p{0.09\linewidth}
  >{\centering\arraybackslash}p{0.13\linewidth}|
  >{\centering\arraybackslash}p{0.09\linewidth}
}
\toprule
 & \makecell{\textbf{Opus}\\\textbf{4.1}}
 & \makecell{\textbf{Gemini}\\\textbf{2.5 Pro}}
 & \textbf{o3}
 & \makecell{\textbf{LLM}\\\textbf{Jury}}
 & \makecell{\textbf{Re-score}\\\textbf{Panels}}
 & $\bm{P}_{\mathrm{severe}}$ \\
\midrule
\textbf{Rate}
 & $5.3\%$
 & $7.4\%$
 & $3.2\%$
 & $4.2\%$
 & $16.7\%$
 & $96.3\%$ 
 \\
\bottomrule
\end{tabular}
\vspace{1em}
\caption{\textbf{Severe error rates} observed relative to primary panels' scores. The probability that the re-score panels have a higher severe error rate than the \lj models ($\bm{P}_{\mathrm{severe}}$) is approximately $96.3\%$.
}\label{tab:severe_rates}
\end{table}

In \textit{Case 2} and \textit{Case 7}, the primary panel noted issues with the CT scan and/or MRI reports, which impacted their diagnoses and scoring. In \textit{Case 4}, the primary panel had access to the full case information, including the age of the patient, which the \lj did not. The primary panel highlighted differences in patient management for a stroke in a younger patient compared to an older patient as the motivation for their high negative treatment risk score. This could not have been considered by the \lj models as patient metadata was not provided in the \lj prompt. Further, diagnoses were paired with ICD-10 codes that do not accommodate this distinction. In \textit{Case 6}, the primary panel gave a low patient safety score because of a perceived omitted tuberculosis diagnosis. However, this was in error as tuberculosis was, in fact, given as a diagnosis. In \textit{Case 9}, the primary panel's motivation for giving a high negative treatment risk focused on errors in differential diagnosis. The \lj scoring reasoning focused on the correct diagnosis, overlooking risks posed by the incorrect differential diagnosis at admission.  

The severe risk assessment errors are good candidates for expert review and suggest improvements to the \lj prompt. For example, the prompt instructed the \lj models to consider the risk to a patient if they are treated according to the reference diagnoses versus the tested diagnoses. The reference and tested diagnoses included the combined list of primary and secondary diagnoses, the differential diagnoses and the clinical reasoning. The prompt could be improved by explicitly stating the relative weightings of these three fields when judging the negative treatment risk. In addition, there are pros and cons when using standardised codes such as ICD-10. \textit{Case 4} illustrates that they can limit the granularity of the diagnosis. Including the patient's age and/or other demographic information in the \lj prompt could mitigate this in some instances.

\section{LLM Jury Model Scoring Stability}\label{sec:score-variablity}

In our exploration of \lj scoring stability, we observed distinct patterns of variability on three randomly selected case diagnoses when inference parameters (default provider settings) were held fixed across $30$ iterations per case. Table~\ref{tab:llm_jury_stability} shows the coefficients of variation (CV) and standard deviations (Std) for each of these cases. The coefficients of variation and standard deviations for the human expert panels were estimated using the scores provided by the primary and re-score panels across 30 cases (Table~\ref{tab:num-dx-eval}).  At the individual case level, there was substantial variation in the computed metrics, likely driven by the integer scoring scale used and varying case complexity. The highest level of internal consistency was observed for o3 across all scores except the differential diagnosis score, maintaining a coefficient of variation and standard deviation that were, on average, lower than those of the other two \lj models. Specifically, o3 achieved a low average relative variability in the patient safety score (CV $0.03$; Std $0.10$), suggesting a highly stable and reliable judgment profile regarding clinical risk assessment. 

In contrast, Gemini 2.5 Pro and Opus 4.1 exhibited higher dispersion in their ratings. Across all \lj models, the diagnosis and clinical reasoning scores showed the greatest fluctuations. Gemini 2.5 Pro had the highest standard deviation in differential diagnosis score ($0.54$), while Opus 4.1 demonstrated the lowest coefficients of variation in this category ($0.08$). These findings indicate that while the \lj is generally stable, specific dimensions, particularly those requiring complex diagnostic logic, are more susceptible to the stochastic elements of model inference. Despite this variability, the average coefficients of variation ($0.03$-$0.21$) suggest that a 30-iteration sampling approach provides a sufficiently converged representation of each model's evaluative consistency.

\section{LLM Jury Scores for Incorrect Diagnosis Detection}\label{sec:bad-cases-selection}

\begin{figure}[htb]
    \centering
    \includegraphics[width=0.8\linewidth]{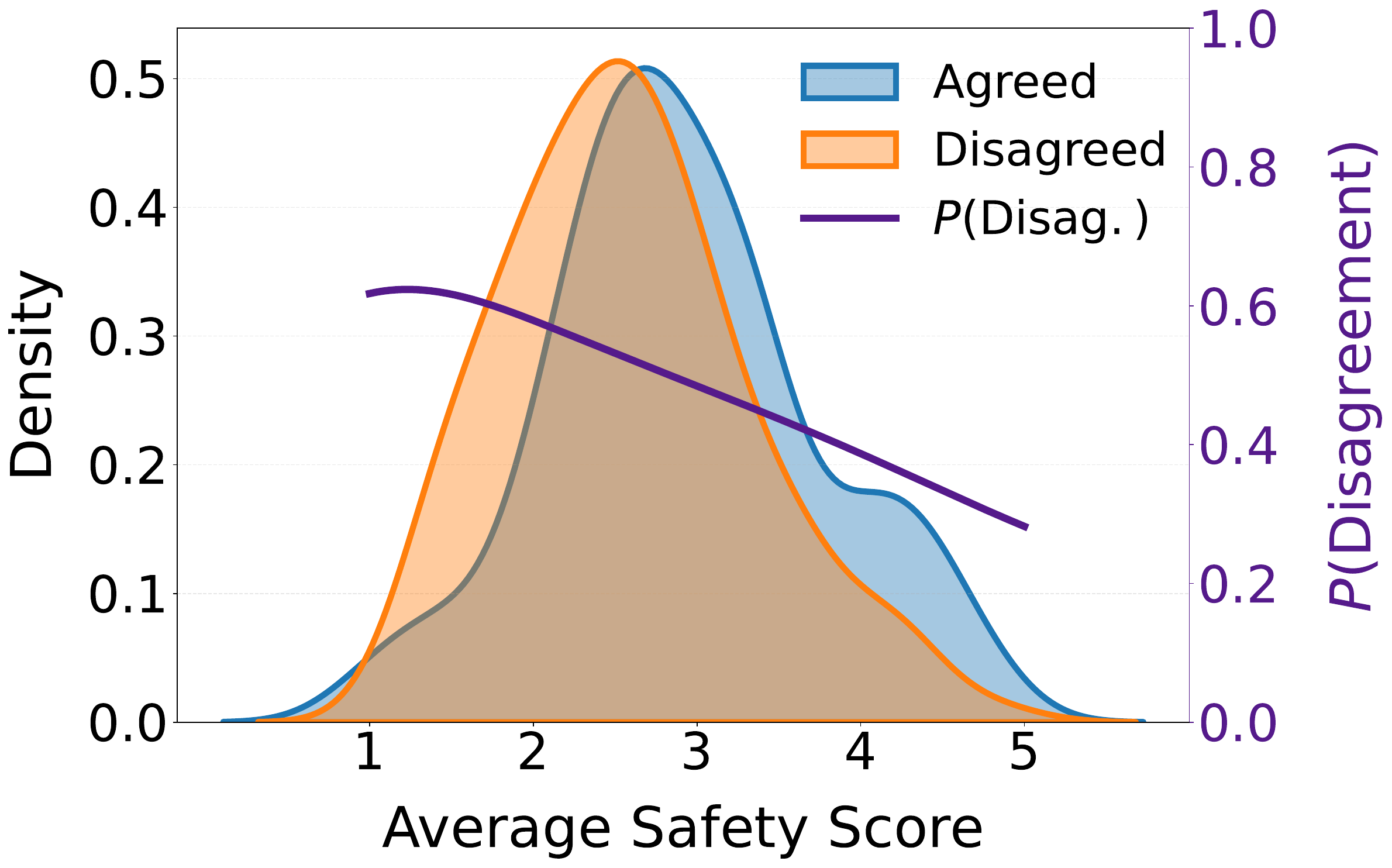}
    \caption{\textbf{Identifying problematic ward diagnoses.} A sample of cases for which the LLM-diagnoses scored poorly on patient safety (Safety), according to the \lj models when the ward diagnoses were used as ground truth, was selected for primary panel review. During review, the panel indicated whether they agreed or disagreed with the ward diagnoses. Shaded regions show the kernel density estimates (KDE) of the mean \LJ patient safety score, using the ward diagnoses as ground truth, subdivided into cases where the primary panels agreed (blue; $n=163$) or disagreed (orange; $n=137$) with the ward. The empirical probability of the primary panel's disagreement with the ward, $P(\text{Disagreement})$ (right axis, purple), is linearly correlated with the \LJ patient safety score. 
    }
    \label{fig:ward-anomaly-detection}
\end{figure}

A novel test of the \lj is its ability to identify cases in which the ward diagnoses were likely incorrect. In this study, $54\%$ of cases reviewed by the panels were selected because all tested AI models scored poorly with the ward diagnoses as ground truth. Specifically, these cases were chosen for primary panel review based on the ranked patient safety scores. During review, the primary panel indicated whether they agreed or disagreed with the ward diagnoses, as mentioned in Section~\ref{sec:data-and-panels}. The probability of primary panel disagreement with the ward diagnosis as a function of the average patient safety score, as scored by the \lj using the ward diagnosis as ground truth, is shown in Figure~\ref{fig:ward-anomaly-detection}. There is a clear negative correlation between the patient safety score and the likelihood of panel disagreement, $P(\text{Disagreement})$. 

\begin{figure}[htb]
    \centering
    \includegraphics[width=0.75\linewidth]{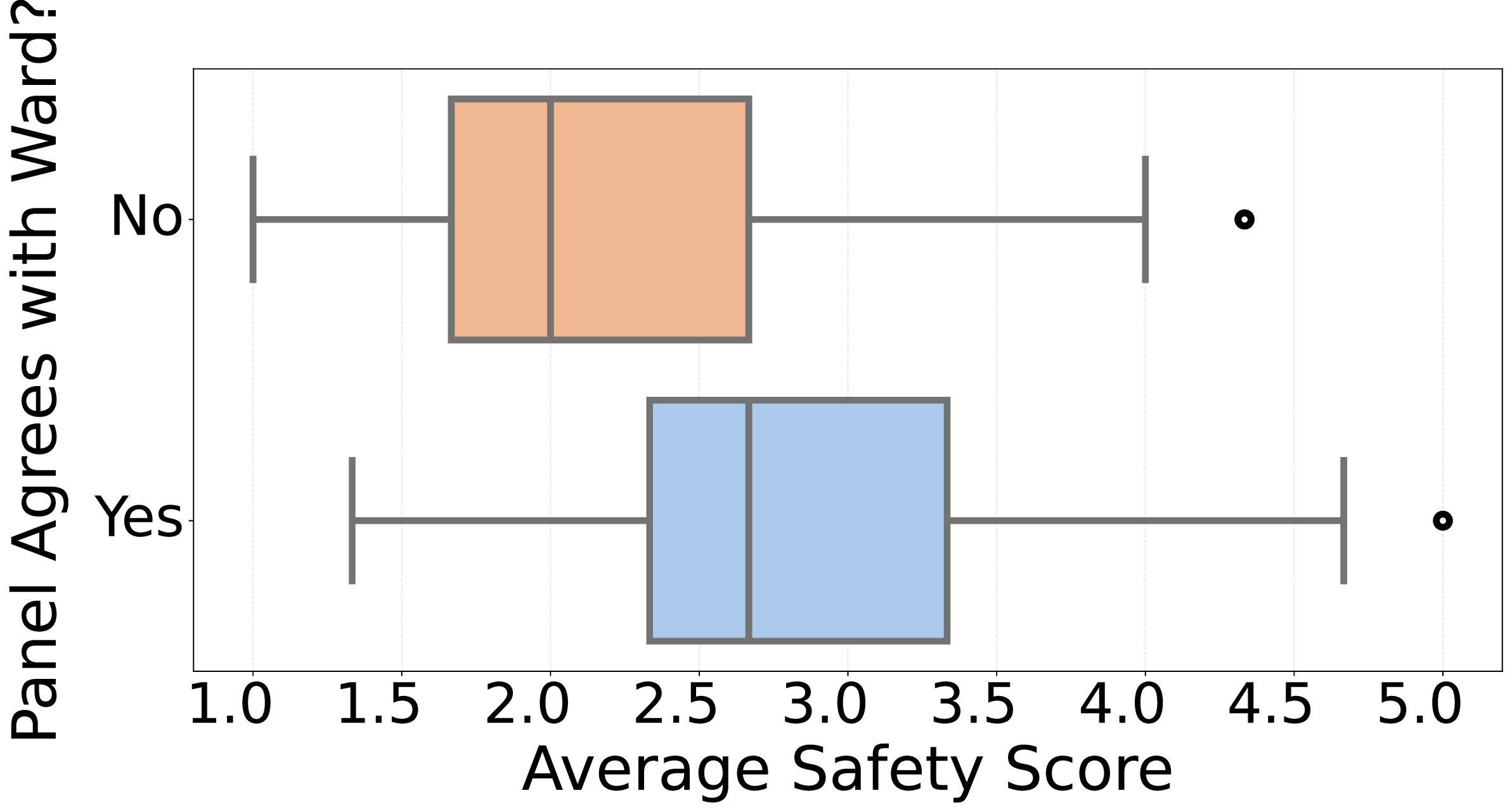}
    \caption{\textbf{Association between diagnostic agreement and LLM Jury patient safety scores.} Distribution of average LLM Jury patient safety scores for ward diagnoses, evaluated using the primary panel's diagnosis as the ground truth. Cases are stratified by whether the primary panel agreed ($n=163$) or disagreed ($n=137$) with the ward diagnosis. On average, panel-ward disagreement corresponds with lower patient safety scores. 
    }
    \label{fig:ward_safety_panel_GT}
\end{figure}

The ward diagnoses were also evaluated by the \lj using the primary panel diagnoses as ground truth. Figure~\ref{fig:ward_safety_panel_GT} compares the ward’s average safety score distributions, as evaluated by the \lj models, for the following disjoint subsets: cases where the primary panel agreed with the ward diagnosis, and cases where they disagreed. There is a clear positive association between primary panel agreement with the ward and higher average patient safety scores. When the panel disagrees with the ward, the distribution of patient safety scores, including the minimum, median, and maximum values, tends to be lower compared to when they agree. Together, Figures~\ref{fig:ward-anomaly-detection} and \ref{fig:ward_safety_panel_GT} provide joint evidence for the quality of the LLM diagnoses and the \lj scoring.

\section{Calibrating LLM Jury Scores to Panels' Scores}\label{sec:cal}

Previously, we observed that \lj models' scores are lower than the expert panels'. To improve agreement with the primary panels, we perform a post hoc calibration of \lj scores using isotonic regression for each \lj model and each score (Dx, DDx, Reasoning, Safety). Isotonic regression enables non-parametric calibration by fitting a monotonic, piecewise-constant function that minimises prediction error while allowing the data to determine the shape of the mapping~\cite{barlow_statistical_1972}. Inputs between the integer training scores (1–5) are mapped via linear interpolation, with outputs constrained to the interval $[1, 5]$. Isotonic regression is well-suited for calibration in this setting because it makes minimal assumptions about the shape of the mapping between \lj scores and the expert panels' adjudications, while enforcing a monotonic relationship that preserves the ordinal structure of the Likert scale. Isotonic regression is particularly appropriate for our use case, where we observed a systematic offset in the \lj scores, as it can flexibly adjust for offset without imposing a linear or parametric form. This results in better-aligned and more interpretable calibrated scores. Moreover, its non-parametric nature allows it to be reliably fit on the relatively small labelled dataset ($n=330$, Table~\ref{tab:num-dx-eval}), while still generalising effectively to the larger unlabelled evaluation sample ($n=3034$, Table~\ref{tab:num-dx-eval}). The calibration dataset is too small to support \LJ fine-tuning. However, comparing calibration to in-context learning should be explored in future work.

To validate the calibration, we performed a 5-fold cross-validation, see Figure~\ref{fig:cal_curves}. Calibration is performed for each \lj model and each score (Dx, DDx, Reasoning and Safety), resulting in 12 isotonic regression models. Table~\ref{tab:summary_calibrated_diff_format} shows the offset, RMSE, Spearman's $\rho$ or Cohen's $\kappa$ after calibration, with the changes shown in brackets. Calibration leads to a significant improvement in offset and RMSE. As expected, no significant changes occur for Spearman's $\rho$. In some cases, Cohen's $\kappa$ improves, while in others it remains comparable to the original performance. 

\begin{figure}[htbp]
    \centering
    \begin{subfigure}[t]{0.47\linewidth}
        \centering
        \includegraphics[width=\linewidth]{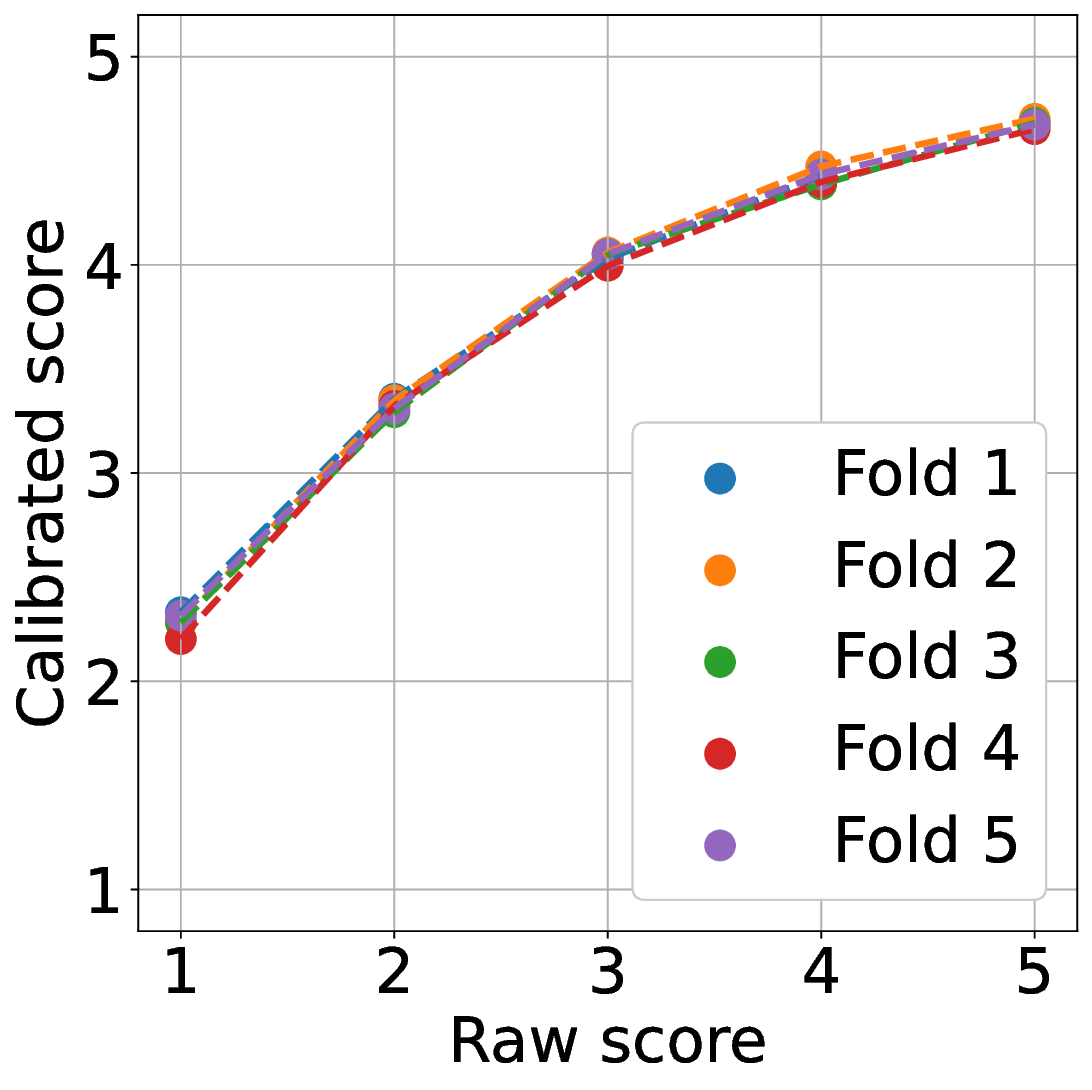}
        \caption{$S_3$ Score}
        \label{fig:cal_s3}
    \end{subfigure}
    \hfill
    \begin{subfigure}[t]{0.47\linewidth}
        \centering
        \includegraphics[width=\linewidth]{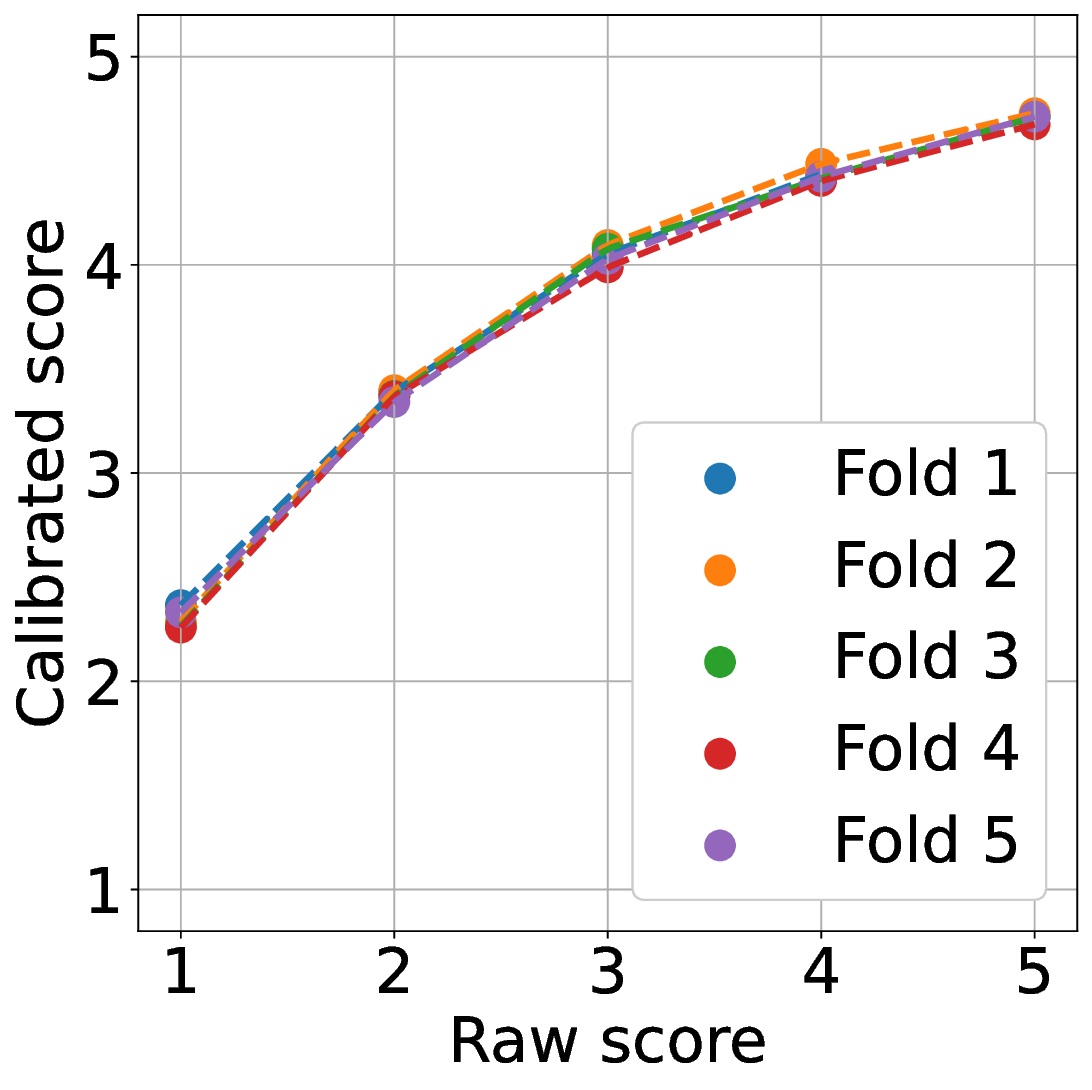}
        \caption{$S_4$ score}
        \label{fig:cal_s4}
    \end{subfigure}
    \caption{\textbf{Calibrated \LJ $\bm{S_3}$ and $\bm{S4}$ scores.} Each \lj model score is calibrated to the expert panel scores using isotonic regression. Calibrated \LJ Dx, DDx, Reasoning and Safety scores are computed as the mean over the corresponding calibrated \lj models' scores. The calibrated \LJ $S_3$ and $S_4$ scores are the weighted sums over the calibrated \LJ scores. Calibration is performed over a 5-fold cross-validation, with line colour indicating different folds. There is very little variation between folds, indicating that the calibration is consistent and generalises well across different subsets. }
    \label{fig:cal_LLM_jury}
\end{figure}

The calibrated \LJ scores are defined as the mean over the calibrated \lj model scores (Figure~\ref{fig:cal_curve_LLM-jury}). We also tested directly calibrating the \LJ scores to the primary panels. However, this did not make a statistically significant change to the performance in terms of offset, RMSE, Spearman's $\rho$ or Cohen's $\kappa$. Similarly, the calibrated $S_3$ and $S_4$ scores introduced and reported in \cite{bassett_multimodal_2026} are recomputed as a weighted sum of the calibrated scores. The $S_3$ and $S_4$ scores use a weighting (0.4, 0.2, 0.0, 0.4) and (0.3, 0.1, 0.3, 0.3) for Dx, DDx, Reasoning and Safety scores, respectively. In particular, the calibrated $S_3$ score provides an overall quality score that is a fair comparison between AI and human diagnostic performance and aligns with the scores provided by the expert human panels. Figure~\ref{fig:cal_LLM_jury} shows the calibration curves for these composite scores across the 5-fold cross-validation averaged over the \lj models (i.e. the calibrated \LJ scores). The calibration is stable across the folds, indicating that it will generalise well. All calibrated scores reported in \cite{bassett_multimodal_2026} were calibrated using Isotonic regression trained on the full calibration sample (n=$330$). 

To confirm that the \LJ provides useful and trustworthy diagnostic evaluations, we compare the \LJ to the primary and re-score panels in terms of diagnostic rankings and error and inter-rater reliability metrics, respectively. In Section~\ref{sec:rankings}, we discuss how the \LJ models' rankings of the diagnosing agents compare to the primary panels' rankings. In Section~\ref{sec:LJ-outperform-rescore}, we compare the performance of the \LJ to the re-score panels in terms of offset, RMSE, Spearman's $\rho$ and Cohen's $\kappa$.  Both analyses assess results before and after calibration. The calibrated \LJ rankings are in good agreement with the primary panel, and the \LJ matches or outperforms the re-score panels across all metrics. 

\subsection{Does the LLM Jury Outperform the Re-score Panels?}\label{sec:LJ-outperform-rescore}

The RMSE and inter-rater reliability metrics, reported in Tables~\ref{tab:offset_rmse} and \ref{tab:rho_kappa}, respectively,  show that for some instances, the \LJ outperforms the human re-score panels. However, these metrics are computed on different samples. To enable a direct comparison, we restrict our analysis to the 30 case diagnoses evaluated by both the re-score panels and the LLM Jury. To estimate the probability that the \LJ outperforms the re-score panels, we compute the bootstrap win rate, i.e. the proportion of case-level bootstrap resamples ($n=1000$) in which the \LJ outperforms the re-score panels for a given metric. Table~\ref{tab:p(A>B)} reports the win rates for offset, RMSE, Spearman’s $\rho$, and Cohen’s $\kappa$, before and after calibration. In addition, we report the mean difference in performance ($\Delta$) to quantify the magnitude of the effect. 

Before calibration, the \LJ matches or outperforms the re-score panels across all scores and metrics, with the exception of offset. For RMSE, Spearman’s $\rho$, and Cohen’s $\kappa$, the win rate for the \LJ exceeds $50\%$ across all scores. The only exception is Cohen’s $\kappa$ for the diagnosis score, where the win rate is $39\%$. However, the mean difference in this case is only $-0.05$, which is negligible given that Cohen's $\kappa$ ranges from $-1$ to $1$. In particular, for Spearman’s $\rho$, the \LJ outperforms the re-score panels across all scores ($75\%$–$100\%$; $\Delta$ $0.1$-$0.4$). The re-score panels substantially outperform the \LJ in terms of offset, consistent with the results in Section~\ref{sec:summary-stats}. The re-score panels achieve win rates of $95\%$–$100\%$, with significant mean offset differences of between $-1$ and $-0.5$.

After calibration, the \LJ matches or outperforms the re-score panels across all scores and metrics, including offset. For offset and RMSE, there is an increase in bootstrap win rate across all scores. The \LJ significantly outperforms the re-score panels in terms of offset and RMSE ($78\%$-$100\%$; $\Delta$ $0.18$-$0.53$), except in the case of offset for the patient safety score, where the \LJ matches the re-score panel. For Spearman's $\rho$, there is little change in bootstrap win rate and $\Delta$.  For Cohen's $\kappa$, the \LJ continues to match the performance of the re-score panel for the diagnosis and patient safety scores, while for the differential diagnosis and clinical reasoning scores, both the bootstrap win rate and $\Delta$ increase after calibration.

\subsection{Do LLM Jury Model Rankings Align with Expert Panels?}\label{sec:rankings}

\begin{figure}[htb]
    \centering
    \begin{subfigure}[t]{0.8\linewidth}
        \centering
        \includegraphics[width=\linewidth]{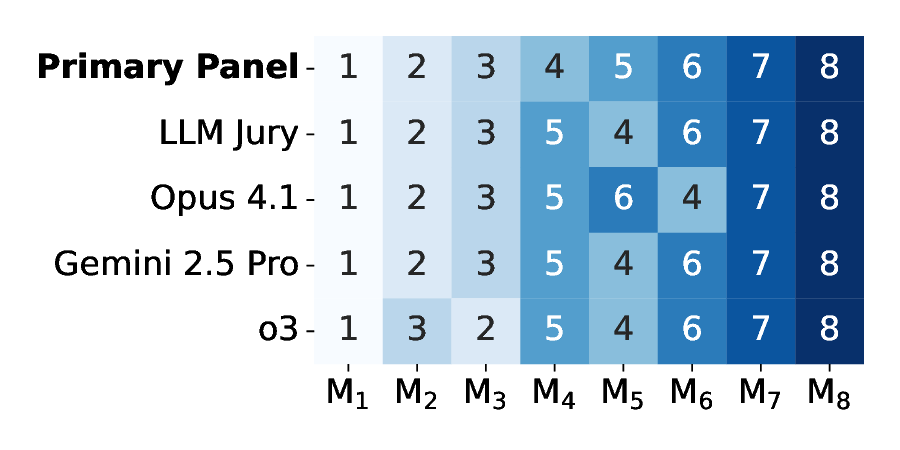}
        \caption{Before calibration}
        \label{fig:ranking_universal_before}
    \end{subfigure}
    \\
    \begin{subfigure}[t]{0.8\linewidth}
        \centering
        \includegraphics[width=\linewidth]{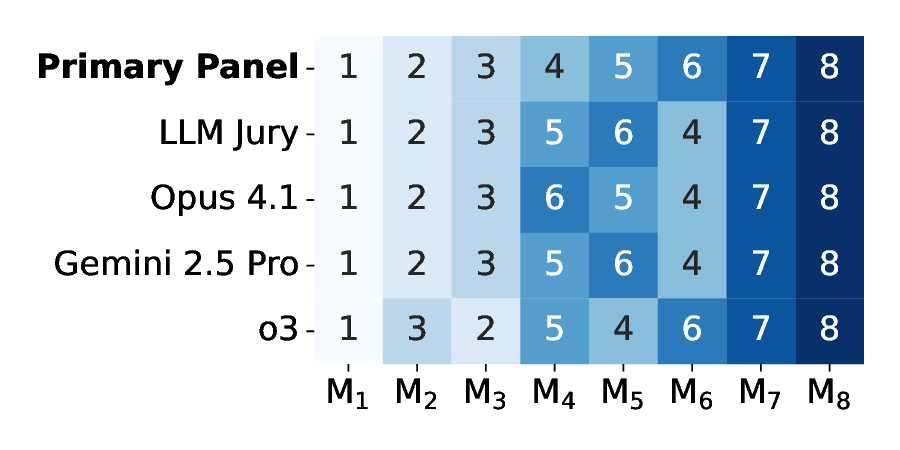}
        \caption{After calibration}
        \label{fig:ranking_universal_after}
    \end{subfigure}    
    \caption{\textbf{Ranking of diagnostic agents.} Comparison of the rankings of the top $8$ diagnostic agents, labelled $M_i$, according to the mean calibrated $S_3$ score provided by each evaluator for the 300 diagnoses reviewed by the primary panel. Text reflects the rankings provided by each evaluator (y-axis), and colours reflect the ranking according to the primary panels. All \lj models have good agreement with the primary panels (Kendall $\tau\geq 0.86$ before and $\geq 0.79$ after calibration). Figure~\ref{fig:ranking_universal_errorbars} shows that o3's ranking of $M_2$ and $M_3$ are interchangeable, and that $M_4$, $M_5$ and $M_6$ are very closely scored by all evaluators. }
    \label{fig:ranking_universal}
\end{figure}

\begin{figure}[hbt]
    \centering
    
    \begin{subfigure}[t]{0.8\linewidth}
        \centering
        \includegraphics[width=\linewidth]{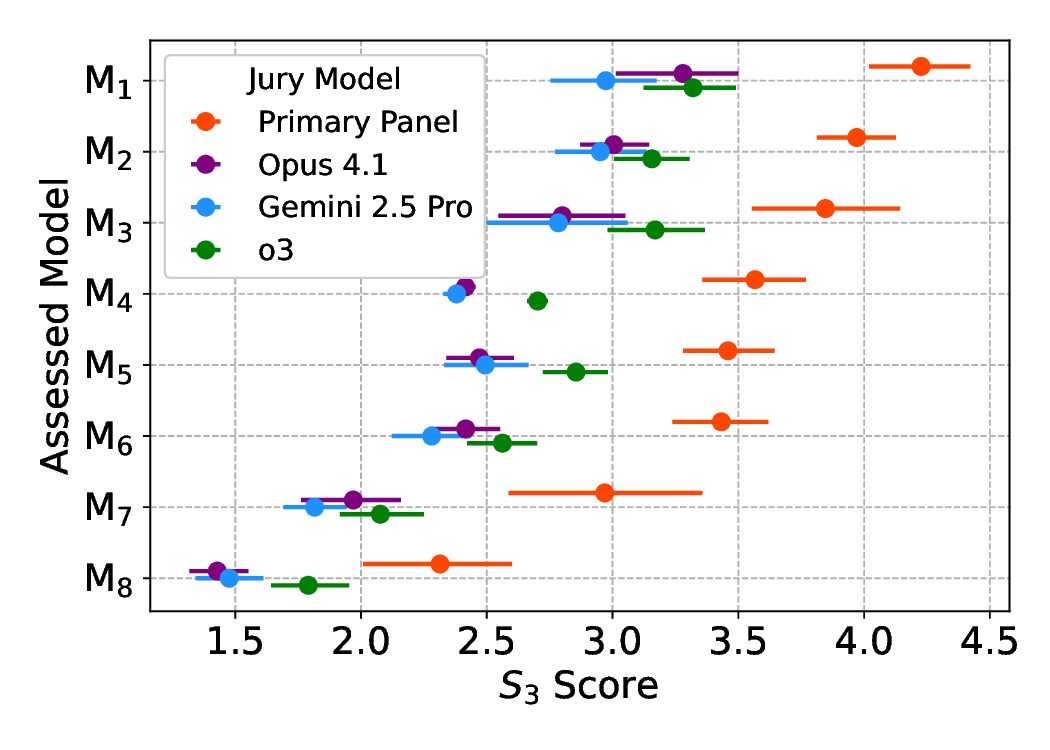}
        \caption{Before calibration}
        \label{fig:ranking_universal_errorbars_before}
    \end{subfigure}
    \\
    \begin{subfigure}[t]{0.8\linewidth}
        \centering
        \includegraphics[width=\linewidth]{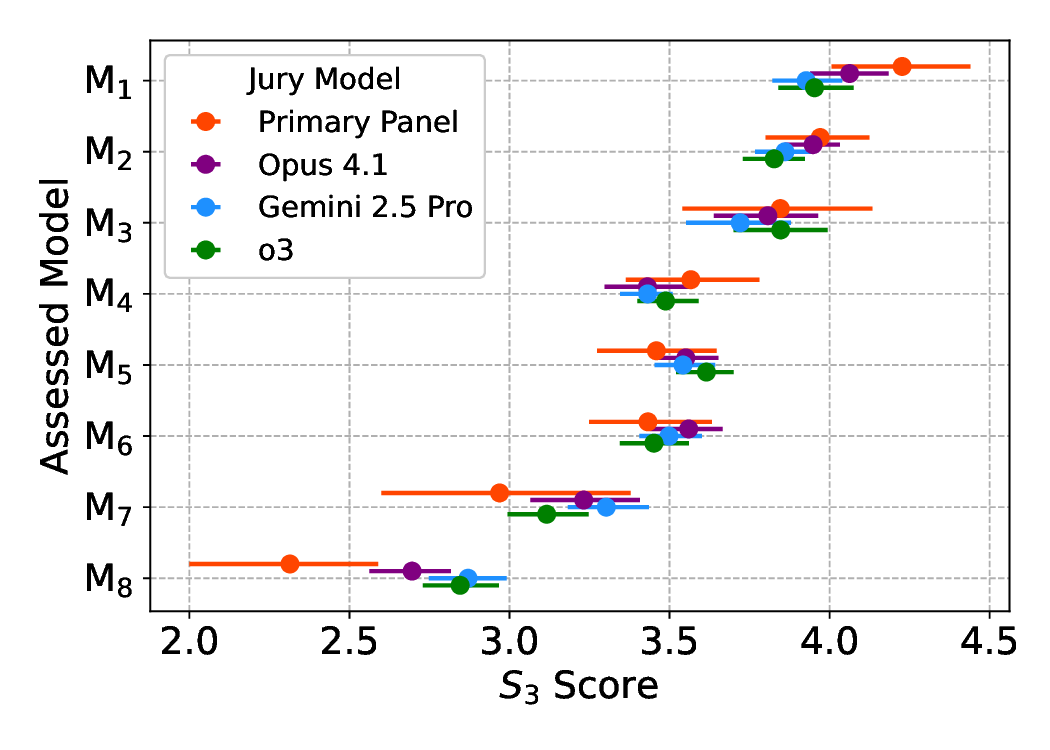}
        \caption{After calibration}
        \label{fig:ranking_universal_errorbars_after}
    \end{subfigure}       
    \caption{\textbf{The $\bm{S_3}$ score distribution.} Mean $S_3$ score for 8 anonymised diagnosing agents ($M_i$) as evaluated by the primary panel and the \lj models, (a) before and (b) after calibration (5-fold cross-validation). The 68\% confidence intervals are computed using bootstrap sampling ($n=1000$). After calibration, there is improved absolute agreement between the \lj models and the primary panels.}
    \label{fig:ranking_universal_errorbars}
\end{figure}

The diagnoses evaluated by the primary panels spanned a variety of models, including the \lj models themselves and models from other providers. Figures~\ref{fig:ranking_universal} and \ref{fig:ranking_universal_errorbars} show the anonymised rankings of $8$ diagnosing agents, labelled $M_i$, based on a `universal quality' score, $S_3$, assigned by the primary panels. The $S_3$ score was computed as a weighted sum of the diagnosis, differential diagnosis, and patient safety scores, weighted 40:20:40, respectively. We only consider the top-8 ranks in this figure due to the small number of evaluations for some models (see Table~\ref{tab:num-dx-eval}), which induces large random fluctuations in ranks. The re-scored panels' scores were also not used to produce rankings due to the small sample size ($n=30$). The subset of models shown was also anonymised since their rankings are not the focus of this paper. Full rankings are reported in \cite{bassett_multimodal_2026}. 

As shown in Figure~\ref{fig:ranking_universal}, there is good agreement between the \LJ rankings of the models $M_i$ and the primary panels' rankings, both before and after calibration. Figure~\ref{fig:ranking_universal_errorbars} shows that the ranking of models $M_2$ and $M_3$ by o3 (green) is interchangeable both before and after calibration, meaning that o3's top 3 rankings effectively match those of the primary panels. Similarly, the group of models $M_4$, $M_5$ and $M_6$ are identically ranked by the \lj models ($M_4$ and $M_6$ ranks are interchangeable for Opus 4.1), while $M_4$, $M_5$, and $M_6$ differ only marginally according to the primary panel. Figure~\ref{fig:ranking_universal_errorbars} also illustrates the improvement in absolute agreement between the \LJ and primary panels, in terms of mean $S_3$ score, after calibration. In summary, the \LJ model rankings align well with the primary panel rankings both before and after calibration, with a Kendall $\tau$ of $0.78$ ($0.82$) before (after) calibration. This provides good evidence that the calibrated \LJ can accurately rank diagnostic agents. 

Across all scores, we found no evidence of scoring bias toward LLM diagnoses from the same provider. \lj models generally ranked their own outputs similarly to how other jury models ranked them. Mixed-effects regression controlling for case-level variability and \lj model effects showed that same-provider evaluations differed by at most $0.041$ points on average, with confidence intervals overlapping zero for all \lj models (Table~\ref{tab:same-provider-bias}). These effects were also small relative to inter-case variability, with same-provider effects ($\leq0.041$) substantially smaller than case-level variance ($\geq0.32$). Each case contributed between $9$ and $30$ same-provider observations, with a mean of $27.3$, providing adequate within-case replication for estimating case-level random effects.

\begin{table}[htbp]
\centering
\small
\renewcommand{\arraystretch}{1.2}
\begin{tabular}{lcccc}
\toprule
\textbf{Score} & 
$\bm{\beta}$ & 
\textbf{Std. Error} & 
\textbf{95\% CI} & 
\textbf{Case Var.} \\
\midrule
Dx            & 0.041 & 0.021 & $[-0.000,\ 0.082]$  & 0.75 \\
DDx      & 0.003 & 0.019 & $[-0.035,\ 0.040]$  & 0.32 \\
Reasoning            & 0.024 & 0.023 & $[-0.021,\ 0.068]$ & 0.87 \\
Safety       & 0.037 & 0.023 & $[-0.008,\ 0.081]$ & 0.66 \\
\bottomrule
\end{tabular}
\vspace{1em}
\caption{Mixed-effects regression estimates of same-provider scoring bias. The coefficient $\beta$ represents the mean change in score when an \lj model evaluates a diagnosis predicted by an LLM model from the same provider. The fact that all $\beta$ values are statistically consistent with zero at 95\% confidence indicates that none of the models treated their own diagnoses differently.  Case variance (Case Var.) corresponds to the estimated random-intercept variance for cases, reflecting inter-case variability.}
\label{tab:same-provider-bias}
\end{table}

\section{Conclusion}

In this study, we provide a detailed assessment of a state-of-the-art \lj for scoring clinical diagnoses and reasoning of both LLM models and routine ward documentation against gold standard references provided by expert human panels of two physicians. We examine complementary approaches for assessing the accuracy, consistency, and reliability of the LLM Jury. Together, these analyses provide a holistic, multi-dimensional evaluation framework for our LLM Jury. As a human reference, we use expert panel scores for diagnoses, reasoning and safety of diagnoses given by a randomly selected LLM on 300 cases. We also have expert human panel re-scoring of 30 of these diagnoses, providing a limited overlap set to test human-human inter-rater reliability. 

We found that the \lj models gave consistently lower scores than the human primary panels. The \lj   
had better inter-rater agreement -- as measured by Cohen $\kappa$ and Spearman $\rho$ -- with each of the two sets of human expert panels (primary and re-score) than the primary-re-score human panels; see Table \ref{tab:rho_kappa}. Importantly, the LLM Jury’s rankings of LLMs for diagnosis closely aligned with those of the primary expert panels, showing that, within our dataset, the LLM Jury can reliably rank diagnostic and reasoning performance.
In our study, the LLM Jury also achieved a significantly lower rate of severe safety-risk errors than the re-score panels.

An independent test of the \lj was provided by using their scores of the LLM model diagnoses, with the routine ward diagnoses as ground truth reference, as a way of selecting high-disagreement cases for expert panel review. We saw good correlation between the \lj scores and the probability that the expert primary panels would disagree with the ward diagnosis.  This enabled targeted expert review and improved evaluation efficiency.

Our analysis demonstrates that post hoc calibration of the \lj scores on the primary panel results using isotonic regression leads to statistically significant improvements in score alignment with panel assessments. Further, among the 30 cases independently evaluated by both the human re-score panels and the LLM Jury, the calibrated LLM Jury matched or outperformed the re-score panels across both accuracy and inter-rater agreement metrics. 
 
Finally, we investigated the question of whether 
LLM jury models favoured diagnoses and reasoning made by models in their same vendor family; e.g. does Gemini 2.5 Pro prefer diagnoses from Gemini models as compared to other models? A mixed-effects regression analysis showed that the results are consistent with none of the \lj models giving preferential treatment (at 95\% CI)  to diagnoses and reasoning provided by their vendor family models. 

Together, these findings, obtained within an LMIC hospital cohort, provide evidence that an \LJ can reliably assess diagnostic accuracy, safety, and clinical reasoning. Future work should evaluate the generalizability of these results across other clinical settings and patient populations.

\section*{Funding} 
This study was funded by the Gates Foundation under Grant INV-081034. While the funder provided useful input to aspects of the study design, they had no authority over the study design; data collection, management, analysis, interpretation, preparation, review, or approval of the manuscript; or the decision to submit for publication. 

\section*{Acknowledgments}
We thank Jess Rees for valuable discussions and/or comments on the draft and Lungile Gabuza, Victor Mngomezulu, Merika Tsitsi, and Sithembiso Velaphi for assistance in conducting the study and Ekram Asefa for software development contributions to the panel platform.

We thank the physicians who participated in the many expert panels: Anu Abraham, Shabbir Alekar, Constance Adams, Nirvana Bharuthram, Tasneem Bux, Dixit Dullabh, Lara Greenstein, Peter Hewsen, Ibraaheem Ismail, Ismail Kalla, Sanjay Lala, Anees Laher, Shannon Leahy, Binu Luke, Siliziwe Lusu, Nhlakanipho Mangeni, Jaishil Manga, Gugulethu Mapurisa, Farzahna Mahomed, David Moore, Pramone Moodley, Aqeela Moosa, Vivendra Naidoo, Sadiya Nanabhay, Bavinash Pillay, Brent Prim, Gary Reubenson, Bianca Rowe, Kebashni Thandrayen, Ismail Tickley and Nicole Van Wyk. 

We acknowledge Lungile Taye, Mahtaab Khan, Tamara Romanini, Christopher Stavrou and Cameron Fisher for assistance in case abstraction.

The authors acknowledge that a version of this work was presented at Artificial Intelligence in Medicine~\cite{Rouillard2026LLMJury}. We thank the anonymous reviewers of Artificial Intelligence in Medicine for their feedback.





\printbibliography

\appendix

\counterwithin{figure}{section}
\counterwithin{table}{section}
\section{Additional figures and tables}

\begin{figure*}[htbp]
    \centering
    \includegraphics[width=0.9\linewidth]{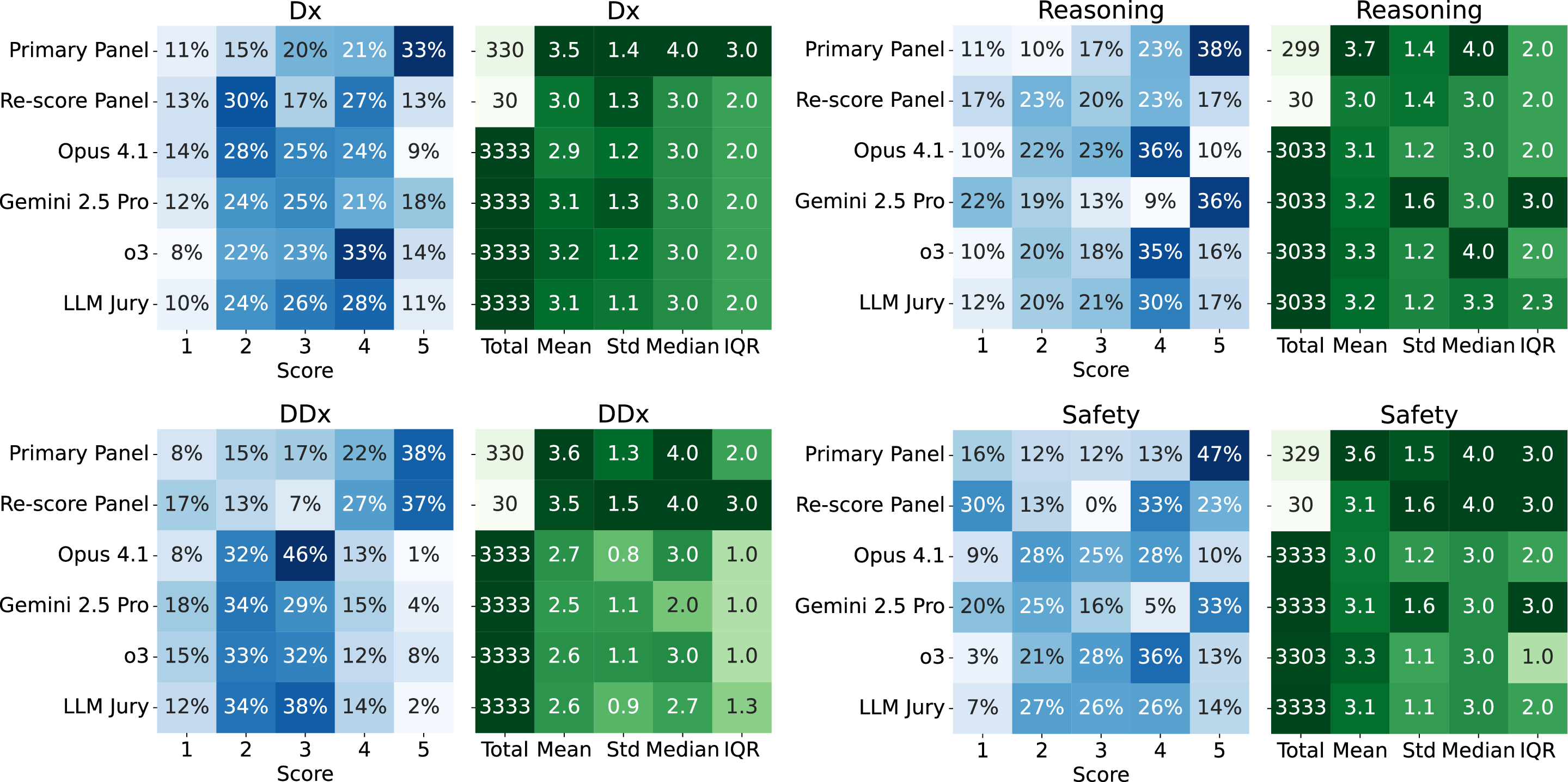}
    \caption{\textbf{Evaluation Summary.} Distribution of scores for each evaluator (blue) and summary statistics (green). The summary statistics include the total number of valid records, the mean, standard deviation (Std), median and interquartile range (IQR) for each evaluator and each score. The total number of diagnoses evaluated by the \lj models is an order of magnitude greater than the number evaluated by the human panels, see Table~\ref{tab:num-dx-eval} for evaluations per diagnostic agent. The ward doctors' clinical reasoning was not recorded, which accounts for the drop in the total number of records for the clinical reasoning (Reasoning) score compared to the other three scores. The mean of each score indicates that the \lj models tend to give lower scores on average than human evaluators. For the differential diagnosis (DDx) score, the \lj has a lower Std and IQR compared to the panels. This is also the case for the patient safety (Safety) score, except for Gemini 2.5 Pro.
    }
    \label{fig:score_distributions}
\end{figure*}

\begin{figure*}[htb]
    \centering
    \includegraphics[width=0.9\linewidth]{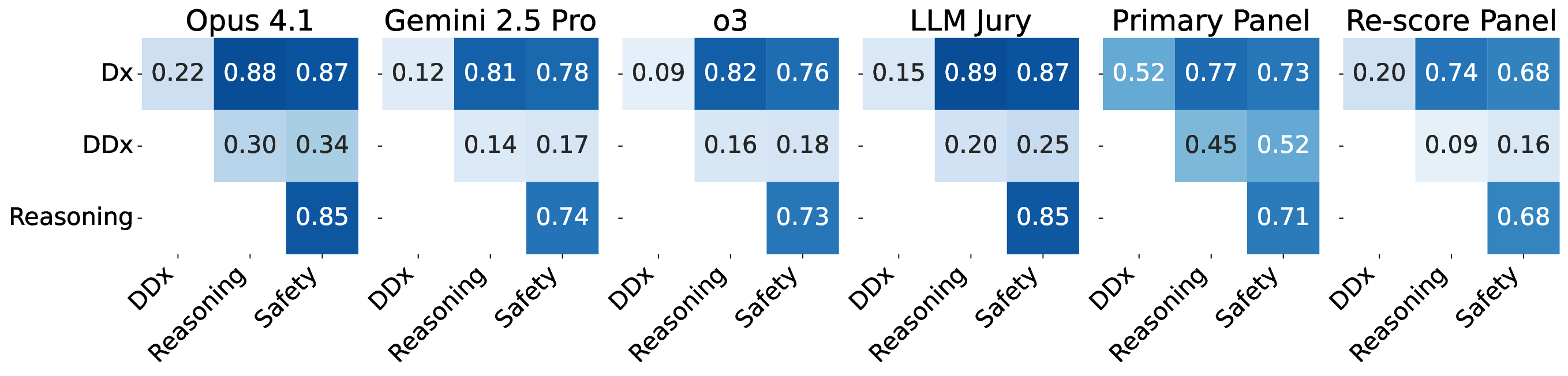}
    \caption{\textbf{Reasoning consistency.} Pairwise correlations between scores assigned by the same evaluator across case diagnoses. Only the upper triangle is shown, as the correlation matrix is symmetric with unit diagonal. For each evaluator, diagnosis (Dx), clinical reasoning (Reasoning), and patient safety (Safety) scores are highly correlated.  The weaker correlation between differential diagnosis (DDx) and the other scores is consistent across all \lj models and mirrors the scoring patterns observed in the human panels.
    }
    \label{fig:inter_score_correlation}
\end{figure*}

\begin{figure*}[htb]
    \centering
    \begin{subfigure}[t]{0.45\linewidth}
        \centering
        \includegraphics[width=\linewidth]{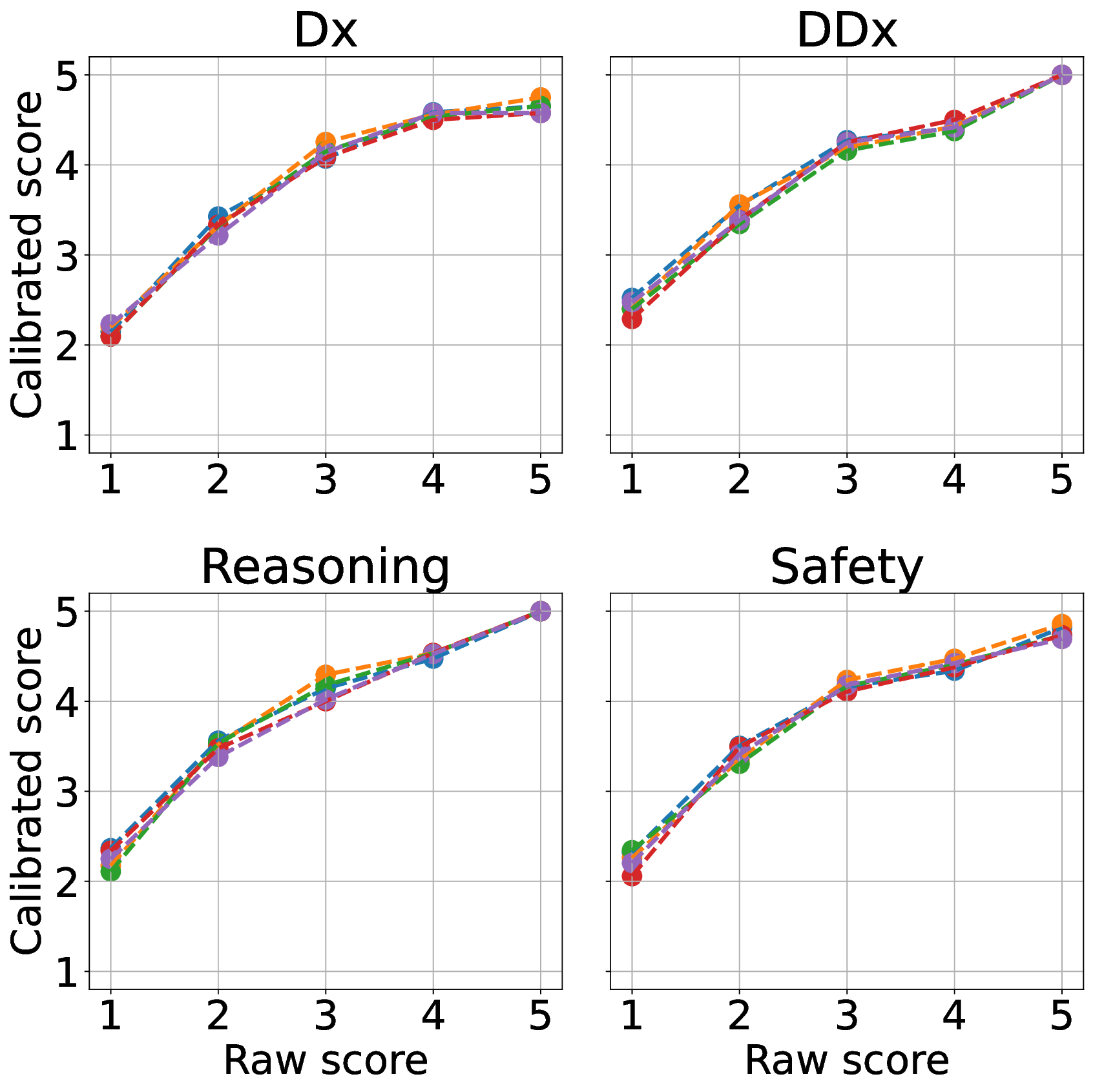}
        \caption{Opus 4.1 
        }
    \end{subfigure}
    \hspace{0.5em}
    \begin{subfigure}[t]{0.45\linewidth}
        \centering
        \includegraphics[width=\linewidth]{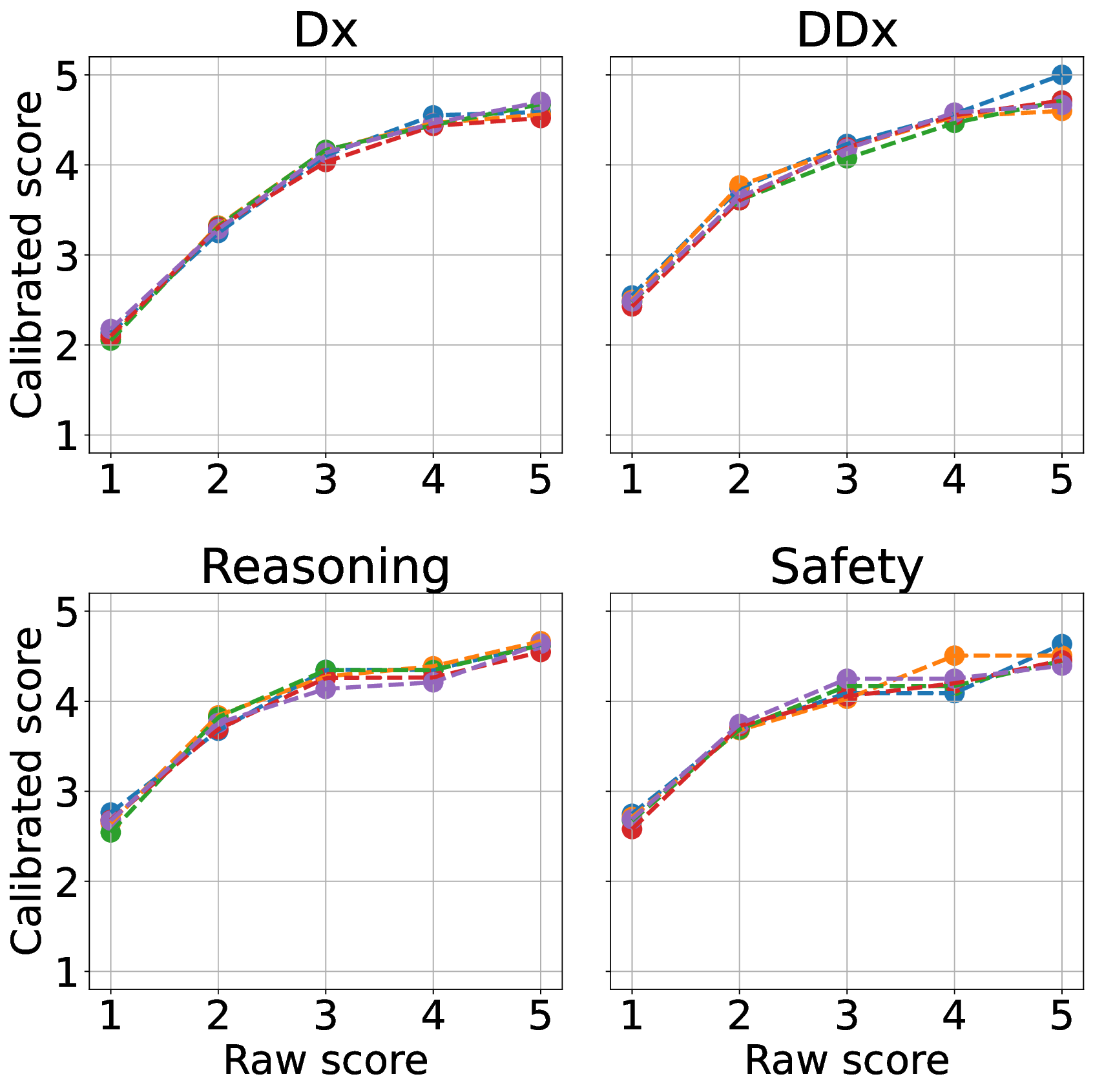}
        \caption{Gemini 2.5 pro 
        }
    \end{subfigure}
    \vspace{0.5em}

    \begin{subfigure}[t]{0.45\linewidth}
        \centering
        \includegraphics[width=\linewidth]{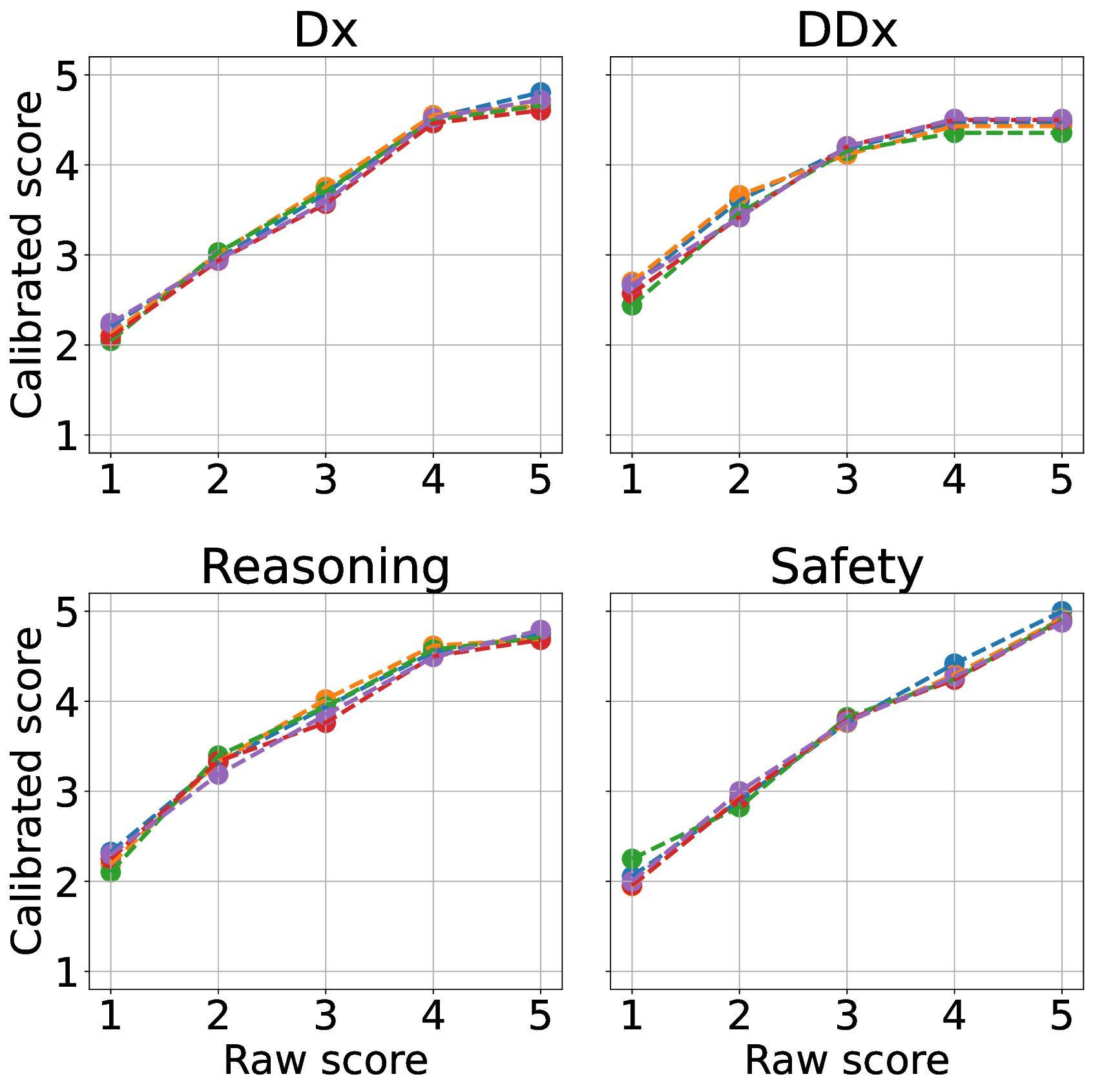}
        \caption{o3 
        }
    \end{subfigure}
    \hspace{0.5em}
    \begin{subfigure}[t]{0.45\linewidth}
        \centering
        \includegraphics[width=\linewidth]{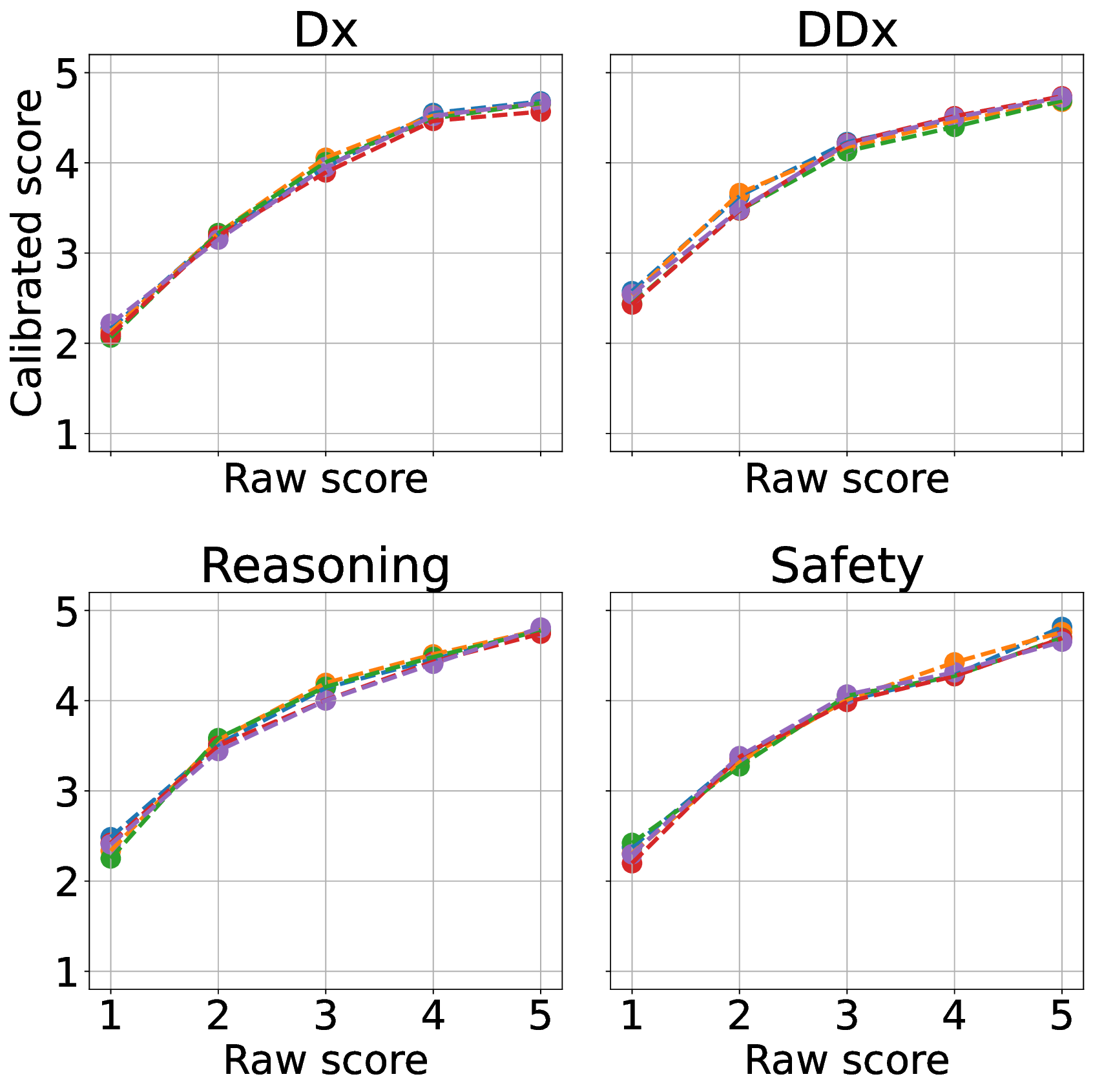}
        \caption{LLM Jury
        }
    \label{fig:cal_curve_LLM-jury}
    \end{subfigure}
    \caption{\textbf{Calibration curves.} Isotonic regression curves for each \lj model, score, and fold in a 5-fold cross-validation. The transformations are highly consistent across folds, suggesting stable calibration. 
    } 
    \label{fig:cal_curves}
\end{figure*}

\begin{table*}[htbp]
\centering
\renewcommand{\arraystretch}{1.2}
\begin{tabular}{llrrrrrrrr}
\hline
\multicolumn{2}{c}{\textbf{Diagnostic agent}} & \multicolumn{5}{c}{\textbf{Calibration Subset - Scored by}} & \multicolumn{3}{c}{\textbf{Evaluation Subset - Scored by}} \\
 \cmidrule(lr){1-2} \cmidrule(lr){3-7} \cmidrule(lr){8-10}
\textbf{Vendor} & \textbf{Name} & \textbf{Primary} & \textbf{Re-score}& \textbf{Opus 4.1 }&\textbf{ Gemini 2.5 Pro} & \textbf{o3 }& \textbf{Opus 4.1} & \textbf{Gemini 2.5 Pro }& \textbf{o3} \\
\hline
- & Ward & 30 & & 30 & 30 & 30 & 300 & 300 & 300 \\

\hline
Anthropic & Claude 4.5 Sonnet &  &  &  &  &  & 197 & 197 & 197 \\
 & Claude 4.1 Opus & 41 & 3 & 41 & 41 & 41 & 195 & 195 & 195 \\
 & Claude 4 Sonnet & 11 &  & 11 & 11 & 11 &  &  &  \\
 & Claude 3.7 Sonnet & 6 &  & 6 & 6 & 6 &  &  &  \\
 & Claude 4 Opus & 5 & 2 & 5 & 5 & 5 &  &  &  \\
 & Claude 3.5 Sonnet & 2 &  & 2 & 2 & 2 &  &  &  \\
 & Claude 3.5 Haiku & 1 &  & 1 & 1 & 1 &  &  &  \\
\hline
Anthropic & All & 66 & 5 & 66 & 66 & 66 & 392 & 392 & 392 \\

\hline
Google & Gemini 2.5 Pro & 54 & 9 & 54 & 54 & 54 & 300 & 300 & 300 \\
 & Gemini 2.5 Flash & 11 & 1 & 11 & 11 & 11 & 300 & 300 & 300 \\
 & Gemini 3 Pro (Preview) &  &  &  &  &  & 298 & 298 & 298 \\
 & Gemini 2.5 Flash (Preview) & 5 &  & 5 & 5 & 5 &  &  &  \\
 & Gemini 2.0 Flash & 4 &  & 4 & 4 & 4 &  &  &  \\
 & Gemini 2.0 Flash-Lite & 3 &  & 3 & 3 & 3 &  &  &  \\
 & Gemini 1.5 Flash & 2 & 1 & 2 & 2 & 2 &  &  &  \\
 & Gemini 2.5 Flash-Lite & 1 &  & 1 & 1 & 1 &  &  &  \\
\hline
Google & All & 80 & 11 & 80 & 80 & 80 & 898 & 898 & 898 \\
\hline
OpenAI & GPT-5.1 &  &  &  &  &  & 288 & 288 & 288 \\
 & o3 & 47 & 5 & 47 & 47 & 47 & 290 & 290 & 290 \\
 & GPT-5 & 32 & 2 & 32 & 32 & 32 &  &  &  \\
 & GPT-4o Mini & 23 & 5 & 23 & 23 & 23 &  &  &  \\
 & o4-mini & 13 & 2 & 13 & 13 & 13 & 291 & 291 & 291 \\
 & GPT-4.1 Mini & 10 &  & 9 & 9 & 9 &  &  &  \\
 & GPT-4.1 & 4 &  & 4 & 4 & 4 &  &  &  \\
 & GPT-4o & 4 &  & 4 & 4 & 4 &  &  &  \\
 & GPT-5 Mini & 3 &  & 3 & 3 & 3 & 287 & 287 & 287 \\
 & GPT-5 Nano & 3 &  & 3 & 3 & 3 &  &  &  \\
\hline
OpenAI & All & 139 & 14 & 138 & 138 & 138 & 1156 & 1156 & 1156 \\
\hline
xAI & Grok 4-1 (Fast Reasoning) &  &  &  &  &  & 288 & 288 & 288 \\
 & Grok 4 (Fast Reasoning) & 15 &  & 15 & 15 & 15 &  &  &  \\
\hline
xAI & All & 15 &  & 15 & 15 & 15 & 288 & 288 & 288 \\
\hline
\textbf{All} & \textbf{All} & \textbf{330} & \textbf{30} & \textbf{329} & \textbf{329} &\textbf{ 329} & \textbf{3034} & \textbf{3034} & \textbf{3034} \\
\hline

\end{tabular}
\vspace{1em}
\caption{\textbf{Number of diagnoses scored.} Summary of the number of case diagnoses, by vendor and name, scored by each evaluator, split by case diagnoses used for score calibration and for final evaluation. The 10 LLMs chosen for final evaluation represented the most up-to-date versions of their variant at the time of data collection. Since the study spanned 12 months, the human panels evaluated a diverse set of LLMs. There is no overlap in case diagnoses between the calibration and evaluation subsets, except for the 30 ward diagnoses that were evaluated by the primary panels and the LLM Jury.  }\label{tab:num-dx-eval}
\end{table*}

\begin{table*}[hbtp]
\centering
\renewcommand{\arraystretch}{1.5}
\begin{tabular}{
p{0.2\columnwidth}
p{0.25\columnwidth} 
    >{\raggedleft\arraybackslash}p{0.18\columnwidth}
    >{\raggedleft\arraybackslash}p{0.18\columnwidth}
    >{\raggedleft\arraybackslash}p{0.18\columnwidth}
    >{\raggedleft\arraybackslash}p{0.18\columnwidth}}
\toprule
\textbf{Metric}
&\textbf{Evaluator} 
& \textbf{Dx}
& \textbf{DDx}
& \textbf{Reasoning}
& \textbf{Safety} \\
\midrule

Offset & Opus 4.1
& \stackci{1.00}{-0.12}{+0.12}
& \stackci{1.30}{-0.13}{+0.14}
& \stackci{1.10}{-0.13}{+0.12}
& \stackci{1.00}{-0.15}{+0.16}
\\

&Gemini 2.5 Pro
& \stackci{0.94}{-0.12}{+0.12}
& \stackci{1.30}{-0.13}{+0.12}
& \stackci{1.10}{-0.17}{+0.16}
& \stackci{1.10}{-0.18}{+0.17}
\\

&o3
& \stackci{0.72}{-0.12}{+0.11}
& \stackci{1.20}{-0.14}{+0.14}
& \stackci{0.86}{-0.13}{+0.13}
& \stackci{0.66}{-0.15}{+0.16}
\\

&LLM Jury
& \stackci{0.89}{-0.11}{+0.12}
& \stackci{1.30}{-0.13}{+0.12}
& \stackci{0.99}{-0.14}{+0.12}
& \stackci{0.93}{-0.15}{+0.14}
\\

&Re-score Panel
& \stackci{\bm{0.40}}{-0.50}{+0.50}
& \stackci{\bm{0.37}}{-0.57}{+0.63}
& \stackci{\bm{0.67}}{-0.53}{+0.53}
& \stackci{\bm{0.10}}{-0.63}{+0.67}
\\

\midrule

RMSE &Opus 4.1
& \stackci{1.50}{-0.10}{+0.09}
& \stackci{1.80}{-0.11}{+0.11}
& \stackci{1.50}{-0.11}{+0.11}
& \stackci{1.80}{-0.11}{+0.11} \\
 
&Gemini 2.5 Pro
& \stackci{1.50}{-0.10}{+0.09}
& \stackci{1.80}{-0.11}{+0.11}
& \stackci{1.80}{-0.13}{+0.13}
& \stackci{2.00}{-0.13}{+0.13} \\

&o3
& \stackci{\bm{1.30}}{-0.10}{+0.09}
& \stackci{1.80}{-0.12}{+0.12}
& \stackci{\bm{1.40}}{-0.12}{+0.11}
& \stackci{\bm{1.60}}{-0.11}{+0.11} \\

&LLM Jury
& \stackci{1.40}{-0.09}{+0.10}
& \stackci{\bm{1.70}}{-0.11}{+0.11}
& \stackci{1.50}{-0.11}{+0.12}
& \stackci{1.70}{-0.12}{+0.12} \\

&Re-score Panel
& \stackci{1.50}{-0.34}{+0.32}
& \stackci{1.80}{-0.52}{+0.45}
& \stackci{1.60}{-0.38}{+0.31}
& \stackci{1.90}{-0.53}{+0.42} \\

\bottomrule
\end{tabular}
\vspace{1em}
\caption{\textbf{Offset and root mean squared error} (RMSE) for each evaluator relative to primary panel scores, averaged over $330$ ($30$) case diagnoses scored by both the primary panels and \lj (primary and re-score panels). The offset is the difference between each evaluator’s score and the primary panel score, where a positive offset indicates that the evaluator's scores are, on average, lower than the primary panel's scores.  The 95\% confidence intervals for the reported means, shown as sub- and superscripts, are computed via bootstrap sampling ($n=1000$). The \LJ refers to scores averaged over the three \lj models. Bold indicates the lowest offset and RMSE for each score. 
}
\label{tab:offset_rmse}
\end{table*}

\begin{table*}[htbp]
\centering
\renewcommand{\arraystretch}{1.5}
\begin{tabular}{
p{0.2\columnwidth}
p{0.25\columnwidth} 
    >{\raggedleft\arraybackslash}p{0.18\columnwidth}
    >{\raggedleft\arraybackslash}p{0.18\columnwidth}
    >{\raggedleft\arraybackslash}p{0.18\columnwidth}
    >{\raggedleft\arraybackslash}p{0.18\columnwidth}}
\toprule
\textbf{Metric}
&\textbf{Evaluator} 
& \textbf{Dx}
& \textbf{DDx}
& \textbf{Reasoning}
& \textbf{Safety} \\
\midrule

Spearman's $\rho$ & Opus 4.1
& \stackci{0.652}{-0.072}{+0.056}
& \stackci{0.463}{-0.098}{+0.097}
& \stackci{0.639}{-0.073}{+0.072}
& \stackci{0.501}{-0.084}{+0.078}
\\

& Gemini 2.5 Pro
& \stackci{0.640}{-0.071}{+0.061}
& \stackci{0.494}{-0.094}{+0.087}
& \stackci{0.590}{-0.079}{+0.078}
& \stackci{0.450}{-0.092}{+0.088}
\\

& o3
& \stackci{0.638}{-0.064}{+0.056}
& \stackci{0.442}{-0.089}{+0.093}
& \stackci{0.628}{-0.079}{+0.066}
& \stackci{0.482}{-0.085}{+0.078}
\\

& LLM Jury
& \stackci{\bm{0.679}}{-0.056}{+0.058}
& \stackci{\bm{0.523}}{-0.090}{+0.084}
& \stackci{\bm{0.656}}{-0.073}{+0.068}
& \stackci{\bm{0.523}}{-0.091}{+0.082}
\\

& Re-score Panel
& \stackci{0.398}{-0.328}{+0.271}
& \stackci{0.113}{-0.359}{+0.374}
& \stackci{0.309}{-0.422}{+0.303}
& \stackci{0.286}{-0.375}{+0.299}
\\

\midrule

Cohen's $\kappa$ & Opus 4.1
& \stackci{0.476}{-0.065}{+0.062}
& \stackci{0.268}{-0.063}{+0.058}
& \stackci{0.467}{-0.070}{+0.068}
& \stackci{0.365}{-0.069}{+0.071}
\\

& Gemini 2.5 Pro
& \stackci{0.487}{-0.069}{+0.061}
& \stackci{\bm{0.293}}{-0.056}{+0.056}
& \stackci{0.423}{-0.071}{+0.075}
& \stackci{0.329}{-0.073}{+0.071}
\\

& o3
& \stackci{\bm{0.541}}{-0.065}{+0.058}
& \stackci{0.285}{-0.068}{+0.068}
& \stackci{\bm{0.514}}{-0.072}{+0.068}
& \stackci{\bm{0.399}}{-0.085}{+0.071}
\\

& LLM Jury
& \stackci{0.438}{-0.053}{+0.056}
& \stackci{0.220}{-0.050}{+0.049}
& \stackci{0.402}{-0.070}{+0.063}
& \stackci{0.325}{-0.070}{+0.062}
\\

& Re-score Panel
& \stackci{0.411}{-0.323}{+0.250}
& \stackci{0.151}{-0.353}{+0.344}
& \stackci{0.309}{-0.350}{+0.281}
& \stackci{0.320}{-0.352}{+0.313}
\\

\bottomrule
\end{tabular}
\vspace{1em}
\caption{\textbf{Spearman's $\rho$ and Cohen's $\kappa$} for each evaluator relative to the primary panels' scores using the $330$ ($30$) case diagnoses scored by both the primary panels and \lj (primary and re-score panels). Higher values indicate better agreement, and bold indicates the best-performing evaluators. The 95\% confidence intervals for the reported metrics, shown as sub- and superscripts, are computed via bootstrap sampling ($n=1000$). Across all scores, the \lj aligns more closely than the re-score panels with the primary panels. The \LJ refers to scores averaged over the three \lj models.
}\label{tab:rho_kappa}
\end{table*}

\begin{table*}[htbp]
\centering\scriptsize
\renewcommand{\arraystretch}{1.2}
\begin{tabular}{
p{0.04\linewidth}
p{0.07\linewidth}
*{3}{>{\centering\arraybackslash}p{0.04\linewidth}} >{\centering\arraybackslash}p{0.04\linewidth}
*{3}{>{\centering\arraybackslash}p{0.04\linewidth}} >{\centering\arraybackslash}p{0.04\linewidth}
*{3}{>{\centering\arraybackslash}p{0.04\linewidth}} >{\centering\arraybackslash}p{0.04\linewidth}
>{\centering\arraybackslash}p{0.06\linewidth}
}
\toprule
\textbf{} & \textbf{} 
& \multicolumn{4}{c}{\textbf{Opus 4.1}} 
& \multicolumn{4}{c}{\textbf{Gemini 2.5 Pro}} 
& \multicolumn{4}{c}{\textbf{o3}} 
& \multicolumn{1}{c}{\textbf{Human}} \\
\cmidrule(lr){3-6} \cmidrule(lr){7-10} \cmidrule(lr){11-14} \cmidrule(lr){15-15}

& 
& 1 & 2 & 3 & Mean
& 1 & 2 & 3 & Mean
& 1 & 2 & 3 & Mean
& Mean \\
\midrule

CV & DDx 
& 0.11 & 0.14 & 0.00 & \textbf{0.08}
& 0.38 & 0.12 & 0.11 & 0.21
& 0.35 & 0.00 & 0.05 & 0.13
& 0.85 \\

 & Dx 
& 0.14 & 0.14 & 0.12 & 0.13
& 0.09 & 0.15 & 0.11 & 0.11
& 0.05 & 0.11 & 0.05 & \textbf{0.07}
& 0.75 \\

& Reasoning 
& 0.15 & 0.11 & 0.07 & 0.11
& 0.20 & 0.04 & 0.08 & 0.11
& 0.00 & 0.07 & 0.08 & \textbf{0.05}
& 0.28 \\

& Safety 
& 0.14 & 0.08 & 0.10 & 0.11
& 0.00 & 0.20 & 0.07 & 0.09
& 0.00 & 0.08 & 0.00 & \textbf{0.03}
& 0.24 \\

\midrule

Std & DDx 
& 0.31 & 0.50 & 0.00 & 0.27
& 0.57 & 0.53 & 0.51 & 0.54
& 0.51 & 0.00 & 0.25 & \textbf{0.25}
& 0.97 \\

& Dx 
& 0.47 & 0.50 & 0.45 & 0.47
& 0.43 & 0.50 & 0.41 & 0.45
& 0.18 & 0.41 & 0.18 & \textbf{0.26}
& 0.92 \\

& Reasoning 
& 0.46 & 0.50 & 0.31 & 0.42
& 0.94 & 0.18 & 0.38 & 0.50
& 0.00 & 0.31 & 0.41 & \textbf{0.24}
& 0.33 \\

& Safety 
& 0.50 & 0.32 & 0.47 & 0.43
& 0.00 & 0.88 & 0.35 & 0.41
& 0.00 & 0.31 & 0.00 & \textbf{0.10}
& 0.28 \\

\bottomrule
\end{tabular}
\vspace{1em}
\caption{\textbf{Score variability.} \lj models' score variability is measured across $30$ repetitions of 3 case diagnoses. Human panel variability is measured across 2 evaluations (primary and re-score panels) of $30$ case diagnoses. Bold indicates the lowest means across \lj models. While the human panel exhibited the highest scoring variability, all \lj models maintained tight consistency, with o3 being the least variable overall.  }
\label{tab:llm_jury_stability}
\end{table*}

\begin{table*}[htbp]
\centering
\renewcommand{\arraystretch}{1.2}
\begin{tabular}{
p{0.25\columnwidth}
p{0.25\columnwidth} 
    >{\raggedleft\arraybackslash}p{0.25\columnwidth}
    >{\raggedleft\arraybackslash}p{0.25\columnwidth}
    >{\raggedleft\arraybackslash}p{0.25\columnwidth}
    >{\raggedleft\arraybackslash}p{0.25\columnwidth}
}
\toprule
\textbf{Metric}
& \textbf{Evaluator}
& \textbf{Dx}
& \textbf{DDx}
& \textbf{Reasoning}
& \textbf{Safety}
\\
\midrule

Offset
& Opus 4.1 & $0.00\ (-1.02)^{*}$ & $0.00\ (-1.27)^{*}$ & $0.00\ (-1.05)^{*}$ & $0.00\ (-1.04)^{*}$ \\
& Gemini 2.5 Pro & $0.00\ (-0.94)^{*}$ & $0.00\ (-1.33)^{*}$ & $0.00\ (-1.08)^{*}$ & $0.00\ (-1.10)^{*}$ \\
& o3 & $0.00\ (-0.72)^{*}$ & $0.00\ (-1.22)^{*}$ & $0.00\ (-0.86)^{*}$ & $0.00\ (-0.66)^{*}$ \\
& LLM Jury 
& $0.00\ (-0.89)^{*}$ 
& $0.00\ (-1.27)^{*}$ 
& $0.00\ (-1.00)^{*}$ 
& $0.00\ (-0.93)^{*}$ \\
\midrule

{RMSE} 
& Opus 4.1 & $1.04\ (-0.46)^{*}$ & $1.18\ (-0.56)^{*}$ & $1.07\ (-0.47)^{*}$ & $1.34\ (-0.41)^{*}$ \\
& Gemini 2.5 Pro & $1.04\ (-0.44)^{*}$ & $1.15\ (-0.65)^{*}$ & $1.13\ (-0.64)^{*}$ & $1.40\ (-0.58)^{*}$ \\
& o3 & $1.06\ (-0.26)^{*}$ & $1.21\ (-0.58)^{*}$ & $1.07\ (-0.37)^{*}$ & $1.36\ (-0.19)^{*}$ \\
& LLM Jury 
& $1.00\ (-0.37)^{*}$ 
& $1.14\ (-0.58)^{*}$ 
& $1.05\ (-0.46)^{*}$ 
& $1.33\ (-0.34)^{*}$ \\
\midrule

{Spearman's $\rho$} 
& Opus 4.1 & $0.60\ (-0.06)\phantom{*}$ & $0.38\ (-0.08)\phantom{*}$ & $0.58\ (-0.06)\phantom{*}$ & $0.45\ (-0.06)\phantom{*}$ \\
& Gemini 2.5 Pro & $0.59\ (-0.04)\phantom{*}$ & $0.43\ (-0.06)\phantom{*}$ & $0.53\ (-0.06)\phantom{*}$ & $0.39\ (-0.06)\phantom{*}$ \\
& o3 & $0.59\ (-0.05)\phantom{*}$ & $0.37\ (-0.07)\phantom{*}$ & $0.57\ (-0.06)\phantom{*}$ & $0.44\ (-0.05)\phantom{*}$ \\
& LLM Jury 
& $0.67\ (-0.01)\phantom{*}$ 
& $0.50\ (-0.02)\phantom{*}$ 
& $0.63\ (-0.02)\phantom{*}$ 
& $0.51\ (-0.01)\phantom{*}$ \\
\midrule

Cohen's $\kappa$
& Opus 4.1 & $0.55\ (+0.07)\phantom{*}$ & $0.39\ (+0.13)^{*}$ & $0.55\ (+0.09)\phantom{*}$ & $0.39\ (+0.02)\phantom{*}$ \\
& Gemini 2.5 Pro & $0.55\ (+0.06)\phantom{*}$ & $0.41\ (+0.11)^{*}$ & $0.46\ (+0.03)\phantom{*}$ & $0.33\ (+0.01)\phantom{*}$ \\
& o3 & $0.46\ (-0.08)\phantom{*}$ & $0.37\ (+0.09)\phantom{*}$ & $0.46\ (-0.06)\phantom{*}$ & $0.33\ (-0.07)\phantom{*}$ \\
& LLM Jury 
& $0.54\ (+0.10)^{*}$ 
& $0.39\ (+0.17)^{*}$ 
& $0.50\ (+0.10)^{*}$ 
& $0.36\ (+0.03)\phantom{*}$ \\
\bottomrule
\end{tabular}
\vspace{1em}
\caption{\textbf{Change in LLM Jury performance after calibration.} Offset, root mean squared error (RMSE), Spearman's $\rho$, and Cohen's $\kappa$ for each \lj model relative to panel scores after calibration. Values in parentheses indicate the difference between the metric before and after calibration, with $^{*}$ denoting statistically significant changes. For offset and RMSE, a negative change is an improvement, while for Spearman's $\rho$ and Cohen's $\kappa$, a positive change is an improvement. Here, statistical significance is defined as the absolute difference exceeding $2\times$ the combined standard error (SE), where the SE is computed using bootstrap sampling ($n=1000$).  Offset, root mean squared error (RMSE), Spearman's $\rho$, and Cohen's $\kappa$ are averaged over $330$ case diagnoses scored by the primary panels. For all metrics, scores and \lj models, the changes in offset and RMSE are statistically significant improvements. Spearman's $\rho$ remains unchanged, while Cohen's $\kappa$ improved in some cases.}
\label{tab:summary_calibrated_diff_format}
\end{table*}

\begin{table*}[htbp]
\centering
\renewcommand{\arraystretch}{1.2}
\begin{tabular}{
p{0.3\columnwidth}
*{4}{>{\raggedleft\arraybackslash}p{0.18\columnwidth}
       >{\raggedleft\arraybackslash}p{0.08\columnwidth}}
}
\toprule
\textbf{Metric}
& \multicolumn{2}{c}{\textbf{Dx}}
& \multicolumn{2}{c}{\textbf{DDx}}
& \multicolumn{2}{c}{\textbf{Reasoning}}
& \multicolumn{2}{c}{\textbf{Safety}}
\\
\cmidrule(lr){2-3}
\cmidrule(lr){4-5}
\cmidrule(lr){6-7}
\cmidrule(lr){8-9}

& {win rate (\%)} & {$\Delta$}
& {win rate (\%)} & {$\Delta$}
& {win rate (\%)} & {$\Delta$}
& {win rate (\%)} & {$\Delta$}
\\
\midrule
Offset (Before) & $0.3$ & $-0.56$
& $0.0$ & $-1.04$
& $0.0$ & $-0.72$
& $5.1$ & $-0.50$
\\
Offset (After) & $\bm{84.1}$ & $+0.25$
& $\bm{78.3}$ & $+0.18$
& $\bm{97.4}$ & $+0.42$
& $43.0$ & $-0.03$
\\\midrule
RMSE (Before) & $73.2$ & $+0.15$
& $56.5$ & $+0.05$
& $\bm{93.6}$ & $+0.35$
& $\bm{88.5}$ & $+0.33$
\\
RMSE (After) & $\bm{94.7}$ & $+0.29$
& $\bm{98.3}$ & $+0.46$
& $\bm{100.0}$ & $+0.53$
& $\bm{94.7}$ & $+0.39$
\\\midrule
Spearman's $\rho$ (Before) & $\bm{75.9}$ & $+0.12$
& $\bm{85.7}$ & $+0.31$
& $\bm{99.7}$ & $+0.40$
& $\bm{90.4}$ & $+0.20$
\\
Spearman's $\rho$ (After) & $\bm{77.2}$ & $+0.12$
& $\bm{84.1}$ & $+0.29$
& $\bm{99.5}$ & $+0.38$
& $\bm{88.3}$ & $+0.19$
\\\midrule
{Cohen's} ${\kappa}$ (Before) & $39.1$ & $-0.05$
& $66.7$ & $+0.09$
& $\bm{86.9}$ & $+0.17$
& $60.5$ & $+0.05$
\\
{Cohen's} ${\kappa}$ (After) & $48.8$ & $-0.01$
& $\bm{78.4}$ & $+0.17$
& $\bm{99.8}$ & $+0.35$
& $58.0$ & $+0.04$
\\
\bottomrule
\end{tabular}
\vspace{1em}
\caption{\textbf{Bootstrap win rates and performance differences.} Win rate denotes the proportion of case-level bootstrap resamples ($n=1000$) in which the \LJ outperforms the re-score panel for a given metric and score. The mean difference in performance ($\Delta$) quantifies the magnitude of the effect. Results are computed on the subset of 30 cases evaluated by both the \LJ and the re-score panels. Before calibration, the \LJ matches or outperforms the re-score panels across all metrics, except for offset. After calibration, the \LJ matches or outperforms the re-score panels across all metrics, including for offset. Bold indicates instances where the bootstrap win rate was $>75\%$. }
\label{tab:p(A>B)}
\end{table*}

\end{document}